\documentclass[format=acmsmall, review=false, screen=true]{acmart}

\usepackage{natbib}
\usepackage{lscape}
\usepackage{rotating}
\usepackage{amsmath,amsfonts,epsfig}
\usepackage{subfigure}
\usepackage{enumitem}
\usepackage[english]{babel}
\usepackage[utf8]{inputenc}
\usepackage[T1]{fontenc}
\usepackage{booktabs}
\usepackage[inoutnumbered,linesnumbered,algoruled,vlined]{algorithm2e}
\usepackage{multirow}
\allowdisplaybreaks
\makeatletter
\DeclareRobustCommand*\cal{\@fontswitch\relax\mathcal}
\makeatother

\newcommand{\veps}{\varepsilon}

\newcommand{\CK}{\mathcal{K}}

\newcommand{\CS}{\mathcal{S}}
\newcommand{\RR}{\mathbb{R}}

\newcommand{\tr}{^{\top}}
\newcommand{\inv}{^{-1}}
\newcommand{\invt}{^{-\top}}

\newif\iftextinsteadofmath

\newcommand{\defv}[2]{
\iftextinsteadofmath
    \expandafter\newcommand\csname#1\endcsname{{\sf #1}}
\else
    \expandafter\newcommand\csname#1\endcsname{#2}
\fi
}

\newcommand{\redefv}[2]{
\iftextinsteadofmath
    \expandafter\renewcommand\csname#1\endcsname{{\sf #1}}
\else
    \expandafter\renewcommand\csname#1\endcsname{#2}
\fi
}

\newcommand{\btab}{\begin{tabbing}}
\newcommand{\etab}{\end{tabbing}}
\newcommand{\ben}{\begin{enumerate}}
\newcommand{\een}{\end{enumerate}}
\newcommand{\be}{\begin{equation}}
\newcommand{\ee}{\end{equation}}
\newcommand{\bea}{\begin{eqnarray}}
\newcommand{\eea}{\end{eqnarray}}
\newcommand{\bean}{\begin{eqnarray*}}
\newcommand{\eean}{\end{eqnarray*}}
\newcommand{\beal}[1]{\begin{alignat}{#1}}
\newcommand{\eeal}{\end{alignat}}
\newcommand{\bealn}[1]{\begin{alignat*}{#1}}
\newcommand{\eealn}{\end{alignat}}
\newcommand{\bd}{\begin{description}}
\newcommand{\ed}{\end{description}}
\newcommand{\bi}{\begin{itemize}}
\newcommand{\ei}{\end{itemize}}
\newcommand{\bc}{\begin{center}}
\newcommand{\ec}{\end{center}}
\newcommand{\bq}{\begin{quotation}}
\newcommand{\eq}{\end{quotation}}
\newcommand{\bco}{}
\newcommand{\bmat}{\left(\begin{matrix}}
\newcommand{\emat}{\end{matrix}\right)}

\newcommand{\ba}[1]{\begin{array}{#1}}
\newcommand{\ea}{\end{array}}

\newcommand{\bfig}{\begin{figure}[htb] \rule{\linewidth}{0.1mm}\\ }
\newcommand{\efig}{ \rule{\linewidth}{0.1mm} \end{figure}}

\newcommand{\bmp}[1]{\begin{minipage}{#1}}
\newcommand{\bmpp}[2]{\begin{minipage}[#1]{#2}}
\newcommand{\emp}{\end{minipage}}

\newcommand{\cI}{\mathcal{I}}
\newcommand{\cJ}{\mathcal{J}}
\newcommand{\cG}{\mathcal{G}}
\newcommand{\cE}{\mathcal{E}}
\newcommand{\balpha}{{\boldsymbol\alpha}}

\newcommand{\Hit}[1]{H^{(#1)}}
\newcommand{\bit}[1]{\beta^{(#1)}}
\newcommand{\xit}[1]{x^{(#1)}}

\newcommand{\nfsx}[2]{\nabla f_{S_{#1}}(x^{(#2)})}

\newcommand{\fx}[1]{f(x^{(#1)})}
\newcommand{\nfx}[1]{\nabla f(x^{(#1)})}

\acmJournal{TKDD}
\acmVolume{0}
\acmNumber{0} 
\acmArticle{0} 
\acmYear{0} 
\acmMonth{0} 
\copyrightyear{0} 
\acmArticleSeq{0}           
\setcopyright{acmlicensed} 
\acmDOI{0000001.0000001}                                                                                                                                            

\newcommand{\BCLR}{{\sc Color-B}\xspace}
\newcommand{\CLR}{{\sc Color}\xspace}
\newcommand{\HW}{{\sc Hogwild}\xspace}
\newcommand{\STR}{{\sc Strata}\xspace}
\newcommand{\BSTR}{{\sc Strata-B}\xspace}

\renewcommand{\thefootnote}{\alph{footnote}}

\acmJournal{TKDD}

\begin{document}

\title[HAMSI]{HAMSI: A Parallel Incremental Optimization Algorithm Using Quadratic
  Approximations for Solving Partially Separable Problems}

\author{Kamer Kaya}
\affiliation{%
  \institution{Sabanc{\i} University}
  \streetaddress{Faculty of Engineering and Natural Sciences}
  \city{Istanbul}
  \postcode{34956}
  \country{Turkey}
}
\author{F\.{i}gen \"{O}ztoprak}
\affiliation{%
  \institution{\.{I}stanbul Bilgi University}
  \streetaddress{Department of Industrial Engineering}
  \city{Istanbul}
  \postcode{34060}
  \country{Turkey}
}
\author{\c{S}evket \.{I}lker B\.{i}rb\.{i}l}
\affiliation{%
  \institution{Sabanc{\i} University}
  \streetaddress{Faculty of Engineering and Natural Sciences}
  \city{Istanbul}
  \postcode{34956}
  \country{Turkey}
}
\author{A. Taylan Cemg\.{i}l}
\affiliation{%
  \institution{Boğaziçi  University}
  \streetaddress{Department of Computer Engineering}
  \city{Istanbul}
  \postcode{34342}
  \country{Turkey}
}
\author{Umut \c{S}\.{i}m\c{s}ekl\.{i}}
\affiliation{%
  \institution{LTCI, T\'{e}l\'{e}com ParisTech, Universit\'{e} Paris-Saclay}
  \streetaddress{LTCI, T\'{e}l\'{e}com ParisTech}
  \city{Paris}
  \postcode{75013}
  \country{France}
}
\author{Nurdan Kuru}
\affiliation{%
  \institution{Sabanc{\i} University}
  \streetaddress{Faculty of Engineering and Natural Sciences}
  \city{Istanbul}
  \postcode{34956}
  \country{Turkey}
}
\author{Hazal Koptagel}
\affiliation{%
  \institution{Bo\u{g}azi\c{c}i  University}
  \streetaddress{Department of Computer Engineering}
  \city{Istanbul}
  \postcode{34342}
  \country{Turkey}
}
\author{M. Kaan \"{O}zt\"{u}rk}
\affiliation{%
  \institution{Sabanc{\i} University}
  \streetaddress{Faculty of Engineering and Natural Sciences}
  \city{Istanbul}
  \postcode{34956}
  \country{Turkey}
}


\begin{abstract}%
  We propose HAMSI (Hessian Approximated Multiple Subsets Iteration),
  which is a provably convergent, second order incremental algorithm
  for solving large-scale partially separable optimization
  problems. The algorithm is based on a local quadratic approximation,
  and hence, allows incorporating curvature information to speed-up
  the convergence. HAMSI is inherently parallel and it scales nicely
  with the number of processors. Combined with techniques for
  effectively utilizing modern parallel computer architectures, we
  illustrate that the proposed method converges more rapidly than a
  parallel stochastic gradient descent when both methods are used to
  solve large-scale matrix factorization problems. This performance
  gain comes only at the expense of using memory that scales linearly
  with the total size of the optimization variables. We conclude that
  HAMSI may be considered as a viable alternative in many large scale
  problems, where first order methods based on variants of stochastic
  gradient descent are applicable.
\end{abstract}




\maketitle

\renewcommand{\shortauthors}{Kaya et al.}

\clearpage

\section{Introduction}
 \renewcommand{\thefootnote}{\arabic{footnote}}

 A vast variety of problems in machine learning are framed as
 unconstrained optimization problems of the form
 \begin{equation}
   \label{eq:objgeneral}
   \min_{x} \sum_{i \in \cI} f_i(x),
 \end{equation}
 where $x$ is a vector of parameters and $f_i$ are a collection of
 functions that implicitly depend on an observed data set. Formally,
 each index $i$ in the set $\cI$ corresponds to a single data item,
 and the number of additive terms in the overall objective
 scales with the size of the available data. 
 A natural approach for
 solving the so called large scale problems (where the set $\cI$ is large) is using parallel computation and a
 divide-and-conquer approach.

 In an optimization context, divide-and-conquer provides an immediate
 solution to the so-called {\it separable} problems. One particular
 toy example of a separable problem is the following:
 \begin{equation}
   \label{eq:objseparable1}
   \min_{x} \{f_1(x_1, x_2) + f_2(x_3) + f_3(x_4, x_5, x_6)\}.
 \end{equation}
 Here, the parameter vector $x = (x_1,\cdots,x_6)$ is partitioned into
 $3$ blocks, denoted by $\balpha_1 = \{1,2\}$, $\balpha_2 = \{3\}$ and
 $\balpha_3 = \{4,5,6\}$ and each objective function $f_i$ depends
 only on $x_{\balpha_i}$ for all $i \in \{1,2,3\}$. We denote the
 parameter index set as $\cJ$ and the function index set as $\cI$; in
 this example, $\cJ = \{1,\cdots, 6\}$ and $\cI = \{1,2,3\}$. As
 individual terms are additive, and $\balpha_i$ are mutually disjoint,
 each subproblem $\min_{x} f_i(x_{\balpha_i})$ can be solved in
 parallel and the results can be combined. 

\begin{figure}[htbp]
\centering
\includegraphics[width=0.45\columnwidth]{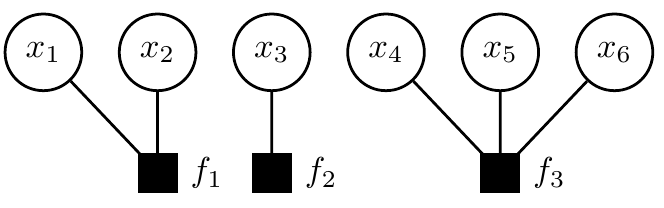} \hspace{0.5cm}
\includegraphics[width=0.45\columnwidth]{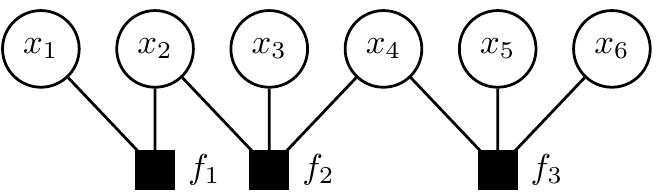}\hfill
\caption{A bipartite graph (factor graph) representation of the two
  toy problems described in the text. (Left) For the separable
  problem, the graph is disconnected and each subgraph corresponds to
  an independent problem and can be optimized in parallel. (Right) For
  partially separable problems, the graph is connected. We can render
  the graph disconnected by removing some function terms (black
  squares).}
\label{fig:fg-example}
\end{figure}

 However, there are many
 optimization problems of interest that are not separable, primary examples being various
 matrix decomposition or regression problems. Fortunately, such
 problems still have some inherent structure that can be exploited for
 parallel computation. As an example, consider a slightly different
 toy example
\begin{equation}
\label{eq:objseparable2}
\min_{x} \{f_1(x_1, x_2) + f_2(x_2, x_3, x_4) + f_3(x_4, x_5, x_6)\},
\end{equation}
where the parameter vector $x$ is again partitioned into $3$ blocks,
$\balpha_1 = \{1,2\}$, $\balpha_2 = \{2,3,4\}$ and
$\balpha_3 = \{4,5,6\}$. Here, the blocks $\balpha_i$ are no longer
mutually disjoint but have relatively small intersections; for
example variable $x_2$ is shared among terms $f_1$ and $f_2$, while
$x_4$ is shared between $f_2$ and $f_3$. We refer to such an objective
function as {\it partially separable}.  Such problems can be
conveniently visualized using a bipartite graph $\cG$ -- also widely
known as a \emph{factor graph} \citep{kschischang2001factor}, albeit
in the context of probabilistic inference. Here, the optimization
variables $x_j$ and the individual objective function terms $f_i$
correspond to nodes, and $\balpha_i$ sets refer to the existing edges
(see Figure~\ref{fig:fg-example}). The graph represents the relation
$j \in \balpha_i$; formally, we have $\cG = (\cI, \cJ, \cE)$ with the
vertex sets $\cI, \cJ$ and the edge set $\cE$ such that for $i\in \cI$
and $j\in \cJ$, we have $\{i, j\} \in \cE$ when $j \in \balpha_i$.

\subsection{Problem Statement}
In the subsequent part of this paper, we shall focus on the following
optimization problem:
\begin{equation} 
\label{eq:probdef}
\min_{x \in \RR^{|\cJ|}} \sum_{i\in \cI} f_i(x_{\balpha_i}),
\end{equation}
where each term $f_i:\RR^{|\cJ|} \mapsto \RR$ for $i\in \cI$ of the
overall objective function $f$ is a twice continuously differentiable
function. The function index set $\cI \equiv \{1,2,\cdots, | \cI | \}$
has typically a very large cardinality $| \cI |$, however, each term
$f_i$ depends only on a small subset of the elements of $x$; that is,
$f_i(x) \equiv f_i(x_{\balpha_i})$. Here $\balpha_i$ are index sets such
that for all $i \in \cI$, $\balpha_i \subseteq \cJ$ where the function
index set $\cJ \equiv \{1,2,\cdots, |\cJ| \}$.  Each singleton
$j \in \cJ$ corresponds to a unique component of vector $x$, denoted
as $x_j$. Thus, if $\balpha = \{j_1, j_2,\cdots, j_A\}$, we have a
vector $x_\balpha = (x_{j_1}, x_{j_1}, \cdots, x_{j_A})$.

There are various strategies for the optimization of partially
separable objective functions. One approach is based on taking
alternating steps, where one can select each coordinate $x_j$ in turns
while regarding the remaining variables $x_{-j}$ as constant, and at
each step solves a smaller optimization problem
$\min_{x_j} f(x_j, x_{-j})$. Alternating least squares, coordinate
descent, or various message passing algorithms are examples of
alternating approaches.

\begin{figure}[t]
\centering
\includegraphics[width=0.45\columnwidth]{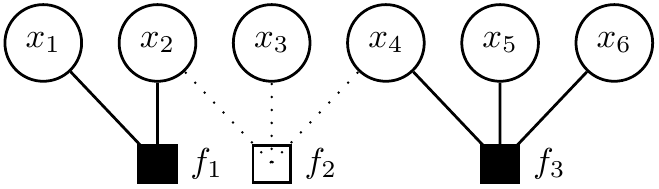} \hspace{0.5cm}
\includegraphics[width=0.45\columnwidth]{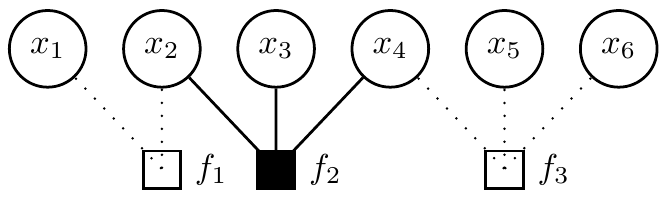}\hfill
\caption{Incremental optimization of a partially separable objective
  by two proxy objectives $f_1(x)+f_3(x)$ and $f_2(x)$. By only
  choosing a subset of the function terms (denoted by black squares)
  and omitting others (white squares) at each iteration we move
  towards the solution of the proxy objective. If the function terms
  are carefully selected, each proxy objective admits parallel
  optimization and by suitably damping the iterations, an approximate
  solution to the true solution can be obtained. We develop a general
  second order method for this general schema.}
\label{fig:fg-example2}
\end{figure}

In this paper, we shall focus on a different strategy, where we select
a collection of terms $f_i$ rather than individual variables
$x_i$. Such methods are known under the name {\it incremental
  methods}; \textit{cf.} \citep{bertsekas:2011}. At each iteration $\tau$, an
incremental method selects only a small subset of the function terms
$f_i$. That is, a subset $S^{(\tau)}$ is selected from the function
indices such that $S^{(\tau)} \subset \cI$. Then, the iterative
algorithm takes a step towards the minimum of a proxy objective
function, $\sum_{i\in S^{(\tau)}} f_i(x)$ (see Figure
\ref{fig:fg-example2}). While at each step $\tau$ a different proxy
objective is used and the actual objective is never evaluated,
incremental algorithms still converge to a solution of the overall
problem under mild conditions, with the stochastic gradient descent
(SGD) being perhaps the most prominent example. In this paper, by
careful selection of subsets $S^{(\tau)}$ at each step $\tau$, we
shall keep the proxy objective separable to allow parallel
computation.

\subsection{A Motivating Example}

The rather generic form of the objective function given by
\eqref{eq:probdef} covers various optimization problems arising in
machine learning. We provide here a particular example to demonstrate
the general approach, yet the formulation can be applied to many other
models and to a wide extent of the application areas. It can be easily
checked, that problems such as sparse logical regression, matrix
completion or multilayer perceptron training can be framed as solving
a partially separable objective function of the form
\eqref{eq:probdef} expressable by a sparse factor graph.

\begin{example}
\label{ex:simple}
  Consider the following matrix factorization problem: 
  \bean 
  \min_{x} \left\| \left( \ba{cc}
      y_{1} & y_{2} \\
      y_{3} & y_{4} \\
      y_{5} & y_{6} \ea \right) - \left( \ba{c}
      x_{1} \\
      x_{2} \\
      x_{3} \ea \right) \left( \ba{cc}
      x_{4} & x_{5} \\
      \ea \right) \right\|_F^2,
   \eean
   where $\|\cdot \|_F$ is the Frobenius norm. Then using our
   notation, the objective function becomes \bean \sum_{i \in \cI }
   f_{i}(x_{\balpha_i}) = (y_1 - x_1 x_4)^2 + (y_2 - x_1 x_5)^2 +
   \cdots + (y_6 - x_3 x_5)^2. \eean Clearly, we have
   $\cI = \{1, 2, \cdots, 6\}$ and $\cJ = \{1, 2, \cdots, 5\}$ with the
   subsets $\balpha_1 = \{1, 4\}$, $\balpha_2 = \{1, 5\}$,
   $\balpha_3 = \{2, 4\}$, $\balpha_4 = \{2, 5\}$,
   $\balpha_5 = \{3, 4\}$, and $\balpha_6 = \{3, 5\}$. The
   corresponding bipartite graph is illustrated in Figure
   \ref{fig:fg}, left. 

   For a more general matrix factorization problem, we have an
   observed data matrix $Y \in \mathbb{R}^{K \times N}$ and we wish to
   find two factor matrices $X_1$ and $X_2$, where
   $X_1 \in \mathbb{R}^{K \times L} $ and
   $X_2 \in \mathbb{R}^{L \times N}$ such that \bean \min_{X_1, X_2}
   \| Y - X_1 X_2 \|_F^2.  \eean In elementwise notation we have \bea
   f(x) = \sum_{a,b} (y_{a,b} - \sum_k x_{1,a,k}
   x_{2,k,b})^2. \label{eq:matrix_fact} \eea Letting $i=(a,b)$, we see
   that this objective can be written as \bean f(x) = \sum_{i \in \cI
   } f_{i}(x_{\balpha_i}), \eean where $\balpha_i$ is the set of
   indices that correspond to the row $a$ of $X_1$ and column $b$ of
   $X_2$. When some entries of the matrix $Y$ are unknown, the
   summation in Eq.\ref{eq:matrix_fact} is merely over the observed
   pairs $(a,b)$, leading still to the form of the final objective.
\end{example}

\subsection{Related Literature}
We propose an incremental and parallel algorithm that incorporates
(approximate) curvature information for distributed large-scale
optimization.  Our experience confirms that using second order
information can accelarate convergence even with incremental
gradients. To gather second order information, the inner problems of
our algorithm are modeled by quadratic functions. Similar to
incremental and aggregate
methods~\citep{bertsekas:2011,tseng:1998,solodov:1998,blatt:2007,roux:2012},
our algorithm exploits the structure of the objective function, as
characterized conveniently by the bipartite graph $\cG$, and evaluates
the gradient only for a subset of the component functions at each
iteration, and it chooses the subsets of component functions in a way
that provides separability of the inner problems. This helps to
distribute the computations over multiple processors and enables doing
stepwise computations on subdomains in parallel. Consequently, modern
distributed and multicore computer systems can be easily utilized.

\citet{gemulla2011} have proposed such a distribution scheme for the
matrix factorization problem, which is employed to deal with
large-scale distibuted relational data thanks to its
partially-separable nature \citep{singh08,cichocki09}. They employ SGD
and analyze its convergence properties. They also consider the
distribution of the problem data on a cluster of computers. The idea
of second order incremental methods has also been
investigated. \citet{bertsekas:1996} has introduced such a method
specifically designed for the least squares problem. His proposal is
an incremental version of the Gauss-Newton method. An extension of
this method for general functions has recently been presented by
\citet{gurbuz:2014}. They have shown linear convergence for the method
under strong convexity and \emph{gradient growth}
assumptions. Moreover, their method requires the computation and the
inversion of the exact Hessian matrices of the component
functions. \citet{shamirDane} have proposed a distributed Newton-like
algorithm for convex problems, where the method requires to compute
and invert $|{\cal J}| \times |{\cal J}|$ Hessian matrices that are
local to each computational node. Due to its generic setting, the
method cannot make use of an explicit parameter space decomposition,
hence requires the \emph{entire} parameter vector to be stored in the
memory and communicated at every iteration.  In another study, an
incremental aggregated quasi-Newton algorithm has been proposed, where
the main idea is to update the quadratic model of one component
function at each iteration \citep{sohl-dickstein:2014}.

Another group of recent work is devoted to stochastic quasi-Newton
methods.  The stochastic block BFGS approach proposed by
\citet{gower2016stochastic} does multi-secant updates, and employs
variance-reduced gradients to achieve linear convergence for convex
problems.  In a recent study, \citet{moritz:2015} have proposed stochastic quasi-Newton methods
that can achieve linear convergence rates under convexity assumptions,
using aggregated gradients and variance reduction techniques. Whilst
these methods have been shown to be useful in certain applications,
they are not suitable for parallel computation due to their gradient
aggregation steps. On the other hand, applying a quasi-Newton method
with stochastic (or incremental) gradients is not straightforward as
it may cause a \emph{data consistency} problem, which becomes more
prominent in a parallel computation setting; \textit{cf.} \citep{schra:2007,
  byrd:2014, berahas2016multi}.

The analysis presented in \citep{gurbuz:2014} for second order
incremental methods does not cover our deterministic algorithm as they
assume convexity of the objective function.  Our analysis follows the
lines of the analysis in \cite{solodov:1998} of incremental gradient
algorithms for nonconvex problems, which does not directly apply to
our case since it does not cover the incorporation of second order
information.  In a recent study, \cite{Wangetal:2014} have analyzed a
framework for stochastic quasi-Newton methods, which applies to
nonconvex stochastic optimization problems. Their analysis covers our
stochastic algorithm considering \eqref{eq:objgeneral} as an expected
value expression with a discrete probability distribution.

\vspace{1em}

In this paper, we make the following contributions:
\begin{itemize}
\item We propose a generic algorithm, which incorporates second order
  information. The proposed algorithm can be applied to a large-class
  of convex {\em and} nonconvex problems, such as; matrix-tensor
  factorization, regression, and neural network training.
\item We demonstrate the convergence properties of our algorithm when
  the subset selection step is deterministic. To the best of our
  knowledge, a proof for such a deterministic algorithm has not been
  given before in the machine learning literature.
\item We investigate several shared-memory parallelization techniques
  and propose a simple cache-friendly load balancing heuristic that
  can be used for stratification-based parallel matrix and tensor
  processing. This latter contribution is particularly important in
  real applications, where the pattern of observed entries in a matrix
  or tensor is far from being uniformly distributed.
\item To present a particular application of the generic algorithm, we
  give an implementation based on L-BFGS \citep{byrd:1994} procedure,
  where we pay special attention to the consistency of the updates.
\item We test our algorithm on large-scale matrix factorization
  problems of varying sizes and obtain faster convergence with
  superior solution times than a well-known first order method
  \citep{gemulla2011}.
\item Our implementation is available
  online\footnote{\url{https://github.com/spartensor/hamsi-mf}}, where
  interested readers can reproduce the results with the best
  performing algorithm in this paper. On the same page, we also report
  our results with the well-known MovieLens datasets. Our results are
  among the best ones to this date.
\end{itemize}

\section{Proposed Algorithm: HAMSI}

The gist of our approach is iterating over multiple subsets of data,
where at each iteration a second order method can be employed. To
derive the generic algorithm, let us first express the function index
set $\cI$ by a union of subsets $S_k$ for $k=1\cdots K$ such that 
\[
\cI = S_1 \cup \cdots \cup S_k \cup \cdots \cup S_K. 
\]
The subsets $S_k$ are not necessary mutually disjoint, but we
typically choose them as such. We will refer to this collection of
subsets a {\it cover}, that will be denoted as
$\CS \equiv \{S_1, \cdots, S_K\}$.  Furthermore, each subset $S_k\in \CS$
for $k = 1\cdots K$ is further partitioned into mutually exclusive
blocks $B_k$ as
\[ 
S_k =S_{k,1} \cup \cdots \cup S_{k,b} \cup \cdots \cup S_{k,B_k},
\] 
where $S_{k,b} \cap S_{k,b'} = \emptyset $ for all $k$ and
$b \neq b'$. The example in Figure~\ref{fig:fg-example2} has a
partition
\[
\cI = S_1 \cup S_2 = (S_{1,1} \cup S_{1,2}) \cup S_{2,1} = (\{1\} \cup
\{3\}) \cup \{2\}.
\]

The cover $\CS$ can be chosen in many ways. Once a cover is fixed,
partitioning of each subset $S_k$ is guided by the factor graph, where
the individual blocks correspond to the terms that can be
independently optimized. This can be achieved by choosing the
partitions $S_{k,b}$ such that the set of variable nodes connected to
the function nodes $f_i$ such that $i \in S_{k,b}$ are mutually
disjoint. Naturally, one can keep the number of blocks $B_k$ large to
keep the degree of parallelism as high as possible.

\begin{figure}[t]
\centering
\includegraphics[width=0.3\columnwidth]{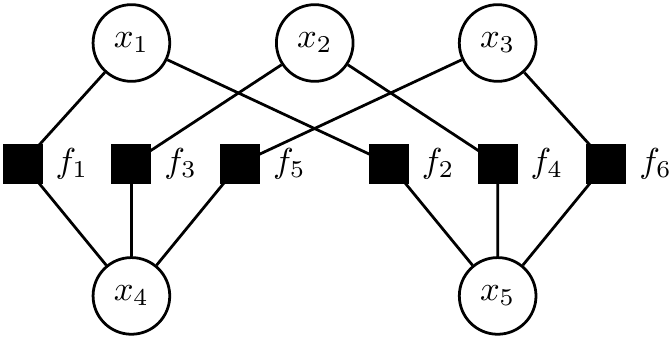} \hfill
\includegraphics[width=0.3\columnwidth]{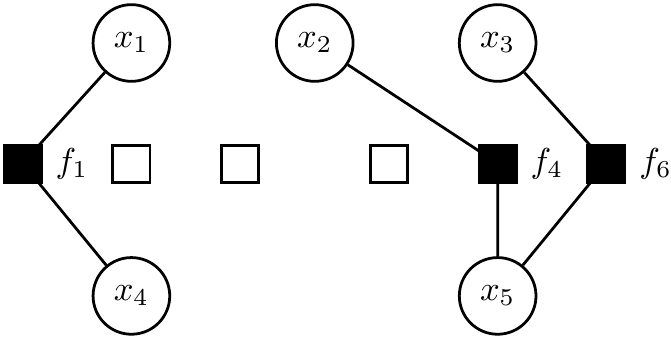}\hfill
\includegraphics[width=0.3\columnwidth]{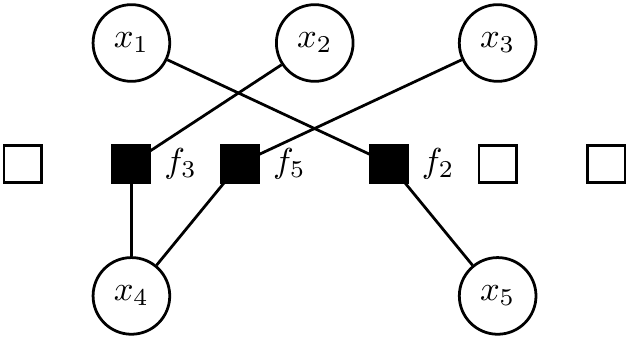}
\caption{(Left) The factor graph of the problem in Example
  \ref{ex:simple}. (Middle and Right) A partitioning of function index
  set as $\cI = \cup_{k=1}^K S_k$ where
  $S_k = S_{k,1} \cup\cdots \cup S_{k,B_k}$ for $k=1\cdots K$. In this
  example $K=2$ and with number of blocks $B_1 = B_2 = 2$. (Middle)
  First subset ($k=1$) with
  $S_1 = S_{1,1} \cup S_{1,2} =\{1\} \cup \{4,6\}$ and
  $\balpha_{1,1}=\{1,4\}$, $\balpha_{1,2}=\{2,3,5\}$, (Right) Second
  subset ($k=2$) with
  $S_2 = S_{2,1} \cup S_{2,2} =\{3,5\} \cup \{2\}$,
  $\balpha_{k,1}=\{2,3,4\}$ and $\balpha_{k,2}=\{1,5\}$. Note that
  each partition renders separable subproblems as the associated
  variable sets are mutually disjoint
  $\balpha_{1,1} \cap \balpha_{1,2}$ and
  $\balpha_{2,1} \cap \balpha_{2,2}$. The resulting algorithm can be
  viewed as alternating between two appropriately chosen matrix
  factorization problems with missing data, each of which can be
  solved efficiently in parallel by exploiting the separability. }
\label{fig:fg}
\end{figure}

The partitions can be understood easily by visualizing the bipartite
graph. For the matrix factorization problem in Example
\ref{ex:simple}, we have illustrated the partitioning in
Figure~\ref{fig:fg}. In a sense, we aim to solve the matrix
factorization problem by approximately solving a sequence of matrix
completion problems where the pattern of missing entries at each stage
is carefully chosen by the algorithm.
Formally, we let 
\[
\cI = \bigcup_{k=1}^K \bigcup_{b=1}^{B_k} S_{k,b},
\]
where the optimization problem with a partially separable objective
function \eqref{eq:probdef} can be written as 
\begin{equation}
\label{eq:obj}
\min_{x \in \RR^{|\cJ|}}  \sum_{k=1}^K \sum_{b=1}^{B_k} \sum_{i\in S_{k,b}} f_i(x_{\balpha_i}).  
\end{equation}
The parallelization within our algorithm relies on the fact that the
objective function in \eqref{eq:obj} is separable over the second
summation indexed by the block index $b$. To achieve this we define
\[
\balpha_{k,b} \equiv \bigcup_{i \in S_{k,b}} \balpha_i \ \
\mbox{ for all } k = 1, \cdots, K; b = 1, \cdots, B_k, 
\]
and require
$\ \balpha_{k,b} \cap \balpha_{k,b'} = \emptyset \ \mbox{ for } b\neq
b' \mbox{ and } \ \bigcup_{b=1}^{B_k} \balpha_{k,b} \subseteq \cJ \
\mbox{ for all } k = 1, \cdots, K$.
The equality $\ \balpha_{k,b} \cap \balpha_{k,b'} = \emptyset$ is
required for the parallel and exact computation of the (partial)
gradients. However, as we describe in Section~\ref{subsec:efficiency},
there exist synchronization-free algorithms in the literature that
accept a small amount of noise in the gradient. For these algorithms,
the parameter sets of the blocks may be overlapping.  In any case we
have,
\[
f_{k,b}(x_{\balpha_{k,b}}) = \sum_{i\in S_{k,b}} f_i(x_{\alpha_i}).
\]
This construction leads to the final form of our optimization problem
that we shall consider in our subsequent discussion:
\begin{equation}
\label{eq:obj_final}
\min_{x \in \RR^{|\cJ|}} \sum_{k=1}^K \sum_{b=1}^{B_k}f_{k,b}(x_{\balpha_{k,b}}).
\end{equation}

\subsection{HAMSI}

Our algorithm uses incremental gradients and incorporates a second
order information into the optimization steps. The second order
information comes from an approximation to the Hessian of the
objective function. As we also work on multiple subsets of $|\cI|$
functions, the algorithm is aptly called Hessian Approximated Multiple
Subsets Iteration (HAMSI). The key idea of the algorithm is using a
local \emph{convex} quadratic approximation 
\begin{equation}
  \label{eq:quad}
  Q(z; \hat{x}, g, H, \beta) \equiv (z-\hat{x})^\top g
  +\frac{1}{2}(z-\hat{x})^\top H(z-\hat{x}) + \frac{1}{2}\beta
  \|z-\hat{x}\|^2 
\end{equation}
for step computation.  Here, $g$ is an incremental gradient, $H$ is
(an approximation to) the Hessian of the objective function. The parameter
$\beta$ is crucial not only to bound the step length but also to control
the oscillation of the incremental steps.

\begin{algorithm}[t]
  \DontPrintSemicolon
  \SetKwInOut{Input}{input}
  \Input{$x$, $H$, $\balpha_{k,b}$ for all
    $k = 1, \cdots, K$; $b = 1, \cdots, B_k$.}
  $t = 1$\;
  ${\CK} = [K, 1, 2, \cdots, K-1, K]$ \; \label{ln:setK}
  \Repeat{convergence or $t$ > max\_epochs}{
    Update $\beta_t$ \ \ (\textit{increasing sequence})\;\label{ln:beta} 
    $\CK \leftarrow$ {\sc SetSchedule}($\CK$) \;
    \For{$\ell = 1, 2, \cdots, K+1$}{  \label{ln:inner}
      $k = {\CK}[\ell]$\; \label{ln:kCl}
      \ForPar{$b=1, 2, \cdots, B_k$}{   \label{ln:update1}
        $x_{\balpha_{k,b}} = \mbox{arg min}_{z} Q(z;\label{ln:quad}
        x_{\balpha_{k,b}}, \nabla f_{k,b}(x_{\balpha_{k,b}}),
        H_{\balpha_{k,b}}, \beta_t)$\; \label{ln:update2}
      } 
    } 
    Evaluate approximate Hessian matrix $H$ at $x$\;
    $t \leftarrow t + 1$\;
  }
  \caption{Hessian Approximated Multiple Subsets Iteration~(HAMSI)}
  \label{algo:hamsi}
\end{algorithm}

Algorithm \ref{algo:hamsi} gives the high-level pseudocode of
HAMSI. In the algorithm, $x$ represents the current solution. At the
$k^{th}$ \textit{inner iteration}, HAMSI uses the functions in $S_k$
(lines \ref{ln:inner}-\ref{ln:update2}). The order of the subsets,
$\CK$, is determined by a {\sc SetSchedule} function which gets the
current order as a parameter, obtains the new order by applying one
cyclic left-shift to the first $K$ elements of $\CK$ and makes sure
that the sequence ends with the same number as the first.  An
illustrative execution of the \textsc{SetSchedule} function for
$K=4$ is as follows:
\begin{center}
  \begin{tabular}{lc}
    Iteration $t$ & $[3, 4, 1, 2, 3] = \text{\sc SetSchedule}([2, 3, 4, 1, 2]) $ \\
    Iteration $t+1$ & $[4, 1, 2, 3, 4] = \text{\sc SetSchedule}([3, 4, 1, 2, 3]) $ \\
    Iteration $t+2$ & $[1, 2, 3, 4, 1] = \text{\sc SetSchedule}([4, 1, 2, 3, 4])$  \\
    $\vdots$ & $\vdots$ 
  \end{tabular}
\end{center}
We note that the subsets are determined only once at the beginning and the \textsc{SetSchedule} function
only reorders a schedule array $\CK$ of size $K+1$ appropriately. 

For each subset, lines
\ref{ln:update1} and~\ref{ln:update2} traverse the corresponding
blocks and update the corresponding parts of the solution, where the
updated part can be denoted as $\bigcup_{b=1}^{B_k} \balpha_{k,b}$. It
is important to note that the corresponding inner loop computes the
blocks of each inner step in parallel and the same (approximate)
matrix $H$ is employed at all inner iterations during the $t^{th}$
cycle. However, the inner iterations use different blocks of $H$
denoted by the submatrix $H_{\balpha_{k,b}}$, where
$H_{\balpha} = \{ H(i,i') : i,i' \in \balpha\}$.  The parameter
$\beta_t$ is also constant during the inner iterations and then it is
updated at the next iteration (line \ref{ln:beta}). The above
description of the algorithm overlooks several important
implementation details; in particular, how to construct the quadratic
approximation in line~\ref{ln:quad} and how to solve the corresponding
subproblems. We shall give an explicit implementation in Section
\ref{sec:impl}, where we exemplify these details.

\subsection{Convergence}
Next, we show the convergence properties of HAMSI when the order of
subset selection is deterministic. Our demonstration follows a similar
construction as given in \citep{mangasarian:1994, solodov:1998}. To
simplify our exposition, we define
\[
\nabla f_{S_k}(x) \equiv \sum_{b=1}^{B_k} \nabla f_{k,b}
(x_{\balpha_{k,b}})
\]
and denote the solution at iteration $t$ by $x^{(t)}$. Since the
approximate Hessian is also evaluated at $x^{(t)}$, we denote it by
$\Hit{t}$. Notice that we update the current solution in lines
\ref{ln:inner}-\ref{ln:update2}. Thus, we denote those inner iterates
by $x^{(t,\ell)}$. Using this notation, the very last inner iterate
$x^{(t,K+1)}$ becomes $x^{(t+1)}$ after line \ref{ln:update2}.

Before presenting our results, let us first list our assumptions:
\begin{enumerate}
\item[A.1] The twice differentiable objective function $f$ is bounded
  below.
\item[A.2] The Hessian matrices for the component functions are uniformly
  bounded at every iteration. That is, for every $S_k$ and $t$, we
  have
  \[
  \| \nabla^2 f_{S_k}(x^{(t)}) \| \leq L,
  \]
  where $L$ is the well-known Lipschitz constant.
\item[A.3] The eigenvalues of the approximation matrices are bounded
  so that
  \[
  (U + \beta_t)^{-1} = U_t \leq \| (\Hit{t} + I\beta_t)^{-1} \| \leq
  M_t = (M + \beta_t)^{-1}
  \]
  holds. Here, $U_t$ and $M_t$ are known constants with
  $0 < M \leq U$, and $I$ denotes the identity matrix.
\item[A.4] The gradient norms are uniformly bounded at every iteration
  $t$. That is, for every $S_k$ we have
  \[
  \|\nfsx{k}{t}\| \leq C,
  \]
  where $C$ is a known constant.
\end{enumerate}

The proof of our convergence theorem depends on two intermediate
results given by two lemmas. The first one establishes a bound on the
difference between the true gradient of a block at $x^{(t)}$ and the
evaluated gradient at the inner iterate $x^{(t,\ell)}$. The second
lemma gives a bound on the error committed by taking incremental steps
at the inner iterates $x^{(t,\ell)}$ instead of the exact Newton step
at $x^{(t)}$. Finally, the convergence theorem uses the boundedness of
the objective function and obtains the desired result by simple
contradiction. The proofs of the two lemmas and the theorem are given
in the Appendix~\ref{sec:appendix}. Note that in line \ref{ln:kCl} of
Algorithm \ref{algo:hamsi}, the subset selection depends on the index
$\ell$. To show this dependence, we shall use the shorthand notation
$S_{[\ell]}$ instead of $S_{\CK[\ell]}$ and define
$\xit{t, 0}\equiv \xit{t}$ to clarify our summations.

\begin{lemma}
\label{lem1}
  In Algorithm \ref{algo:hamsi}, let $x^{(t)}$ be the solution at
  iteration $t$ and $x^{(t, \ell)}$ be the inner iteration
  $\ell$. Then, we have
  \begin{equation}
    \label{eq:lem1}
   \| \nfsx{[\ell]}{t,\ell-1} - \nfsx{[\ell]}{t} \| \leq LM_tC(\ell-1) .
  \end{equation}
\end{lemma}

Next lemma gives a bound on the difference between two consecutive
iterations.

\begin{lemma}
\label{lem2}
Given two consecutive iterations, $t$ and $t+1$ of Algorithm
\ref{algo:hamsi}, we have
  \[
  \| \xit{t+1} - \xit{t} \| \leq \frac{B + C(M+1)}{M + 1}M_t,
  \]
  where $B \equiv \frac{1}{2}L C K(K+1)$.
\end{lemma}

Finally, we present the convergence of the proposed algorithm.

\begin{theorem}
\label{thm:conv1}
Consider the iterates $\xit{t}$ of Algorithm \ref{algo:hamsi}. Suppose
that $\bit{t} \geq 1$ is chosen to satisfy
\begin{equation}
\label{cond_beta}
\sum_{t=1}^{\infty} U_t = \infty \mbox{ and } \sum_{t=1}^{\infty}
M^2_t < \infty.
\end{equation}
Then, 
\[
\underset{t \uparrow \infty}{\lim} \nfx{t} = 0,
\]
and for each accumulation point $x^*$ of the sequence
$\{\xit{t}\}$, we have $\nabla f(x^*) = 0$.
\end{theorem}
Clearly, the simple choice of $\bit{t} = (\eta t)^\gamma$ with
$\eta > 0$ and $\gamma \in (0.5, 1]$ satisfies condition
\eqref{cond_beta}. 

We note for non-convex problems that analyzing the rate of convergence
with Hessian approximations is known to be very difficult;
\textit{cf.}  \citep{shamirDane}. We shall, however, demonstrate the
convergence behavior of HAMSI empirically on large-scale problems in
Section \ref{sec:compstud}.

\section{Partitioning and Parallelization} 
\label{sec:par} 

Our second aim in this paper is to investigate various parallelization
strategies for HAMSI. Naturally, these strategies can also be applied
to other gradient descent based optimization algorithms. As expected,
the computational efficiency of the approach hinges critically on the
choice of independent subproblems (partitioning terms into groups). A
parallel optimization algorithm requires finding a suitable cover
$\CS = \{S_1, \cdots, S_K\}$ of the function index set $\cI$, and a
partitioning of each subset $S_k \in \CS$ into a number of
blocks. Taking the number of synchronization points required for a
correct parallel execution into account and enforcing load balancing,
the cover generation and partitioning problems lead themselves to
various interesting combinatorial optimization problems, such as graph
coloring and balanced stratification. In
Section~\ref{subsec:efficiency}, we describe three parallelization
techniques and two of their variants which obey the generic form of a
partially separable objective function of Equation~\ref{eq:obj_final}.

The first parallelization scheme, \CLR, uses a {\em valid} coloring of
the bipartite graph representation at
Figure~\ref{fig:fg-example}. Given a bipartite graph
$\cG = (\cI, \cJ, \cE)$, the function vertices in $\cI$ are first
colored with $K$ such that two $\cI$ vertices that are adjacent to at
least one $\cJ$ parameter vertex, have different colors. If this is
the case, we say that the coloring is {\em valid}. Once a coloring of
the $\cI$ vertices are given, they can be partitioned into disjoint
subsets with respect to their colors. Clearly, this partitioning
defines a cover $\CS = \{S_1, \cdots, S_K\}$, where each block
$S_{k, b} \subseteq S_k$ contains a single function for all
$k = 1, \cdots, K$ and $b = 1, \cdots, B_k$. Thanks to the validity of
the coloring, all the blocks in the same subset can be processed in
parallel since $\balpha_{k,b} \cap \balpha_{k,b'} = \emptyset$ for all
$1 \leq b < b' \leq B_k$; that is, the parameter sets of the blocks
are disjoint. However, for a parallel, lock-free execution, one still
needs a synchronization point in between $S_k$ and $S_{k'}$, since the
index sets $\balpha_{k,b}$ and $\balpha_{k',b'}$ may overlap for some
$1 \leq b \leq B_k$ and $1 \leq b' \leq B_{k'}$ pair. Hence, the
subsets $S_k$ can be processed in any desired order but not
concurrently.

From the parallelization point of view, there are two desired
properties of a coloring-based execution: (i) the number of
synchronization points should be as small as possible; (ii) the load
distribution among the processors should be balanced. The former is
implied by the formal definition of the graph coloring problem, which
necessitates the minimum number of colors. Unfortunately, the problem
is NP-Complete~\citep{matula_SL} and hard to
approximate~\citep{Zuckerman} for most of its variants. Fortunately,
there exist cheap coloring heuristics in the literature that keep the
number of colors small and can also be employed for the
bipartite-graph coloring problem described
above~\citep{ColPack}. These heuristics have been succesfully used to
compute Hessian and Jacobian matrices and derivatives in
parallel~\citep{Gebremedhin02paralleldistance-k,GMP05,Bozdag10-SISC}. In
this work, we employ a sequential greedy heuristic
from~\cite{ColPack}, which employs a {\em first-fit policy} and visits
the vertices in some order and greedily colors the current vertex with
the first available color (the one with the smallest index) that is
consistent with the previous coloring decisions.

\begin{figure}[htbp]
  \begin{center}
  	\includegraphics[width=0.60\columnwidth]{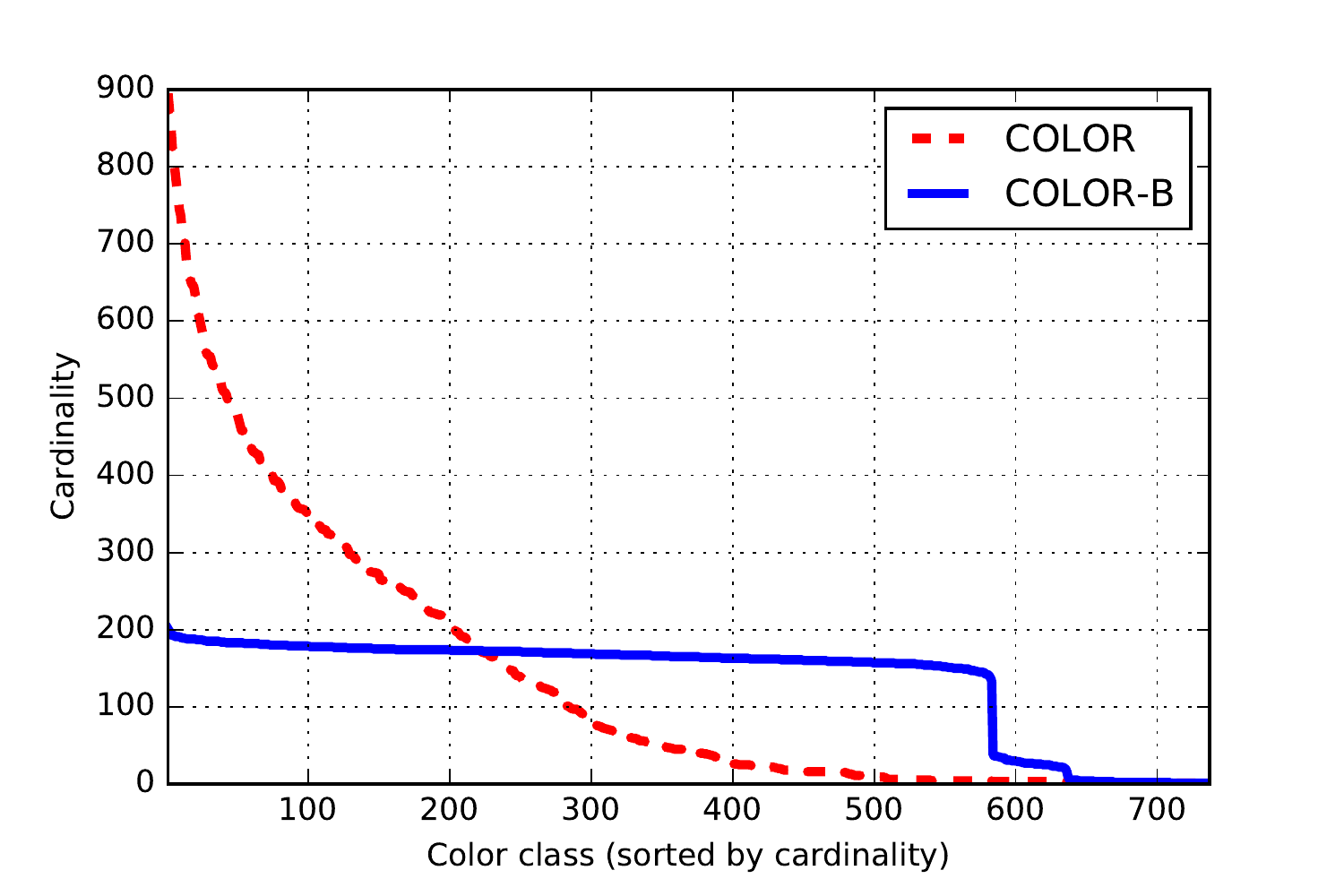}
     \end{center}
  \caption{The number of functions in each color set (sorted with
    respect to their cardinalities) for MovieLens 100K dataset when a
    greedy coloring approach with a first-fit (\CLR) and random
    available color (\BCLR) policy is applied.}
   \label{fig:bcolor}
\end{figure}

Given a coloring, the subsets $S_k$ for $k = 1, \cdots, K$ are well
defined. While processing a subset, it is straightforward to evenly
distribute its blocks to the processors since their parameters are
disjoint and sizes are uniform: in \CLR, the threads process (almost)
the same number of blocks of the same subset. However, in our
preliminary experiments, we observed that the first-fit policy
generates a few very large color sets~(ones with smaller index) and
thousands of color sets~(larger indexed colors) with less than five
functions. In general, when we have more than $B_k$ threads to process
$S_k$, only $B_k$ of the threads will be fed with work. Fortunately,
the small-indexed color sets contain enough work to saturate each
thread in a single, shared memory architecture, and these sets contain
almost all of the functions when combined. Nevertheless, to {\em
  balance} the color set cardinalities, we modify the first-fit policy
and instead of choosing the first available color for a function, we
choose a {\em random available color}.\footnote{An already used random
  color is chosen if possible, otherwise, a new color is chosen.} This
scheme, \BCLR, fills the large-indexed colors better and yields a
balanced function-to-color distribution. Figure~\ref{fig:bcolor}
illustrates this point for the MovieLens 100K
dataset.

As our experiments show, the minimum number of colors for a valid
coloring can be in the order of thousands in practice and hence, the
number of subsets, $K$, cannot be set to a very small number. To solve
this problem, for \CLR and \BCLR, we have packed the color sets into
$K$ bins such that each bin has almost an equal number of
functions. Given $K$, the packing is performed in a greedy fashion by
applying a first-fit policy and putting a color set into the
first available bin having less than $|\cI| / K$ functions. After the
packing, the $k$th bin is used to form the subset $S_k$.

The second parallelization scheme we employed, \HW, does not consider
synchronization at all \citep{Hogwild}. Hence, all the synchronization
overhead can be avoided at the cost of computing a potentially
inaccurate (partial) gradient. In this scheme, the functions in $\cI$
are randomly distributed into $K$ subsets, where each subset $S_k$ for
$k = 1, \cdots, K$ contains (almost) the same number of functions. As
in \CLR, a block is defined as a single-function. In this scheme, the
parameter sets of the blocks in a subset can be overlapping, and when
they are processed in parallel, some entries of the gradient may be
computed incorrectly due to potential race conditions. Considering the
large size of the parameter set in practice and the limit on the
number of cores on a single machine, \HW assumes that there will be no
error in the gradient, and even there is, its impact will be
insignificant.

\begin{figure}[t]
  \begin{center}
    \subfigure[\STR]{\includegraphics[width=0.46\columnwidth]{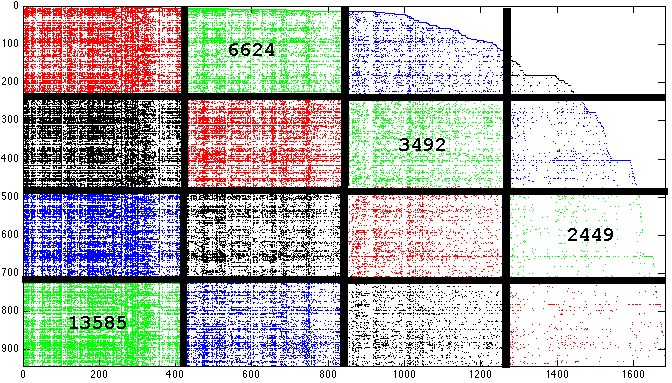} 
      \label{fig:strat}
    }
    \subfigure[\BSTR]{\includegraphics[width=0.46\columnwidth]{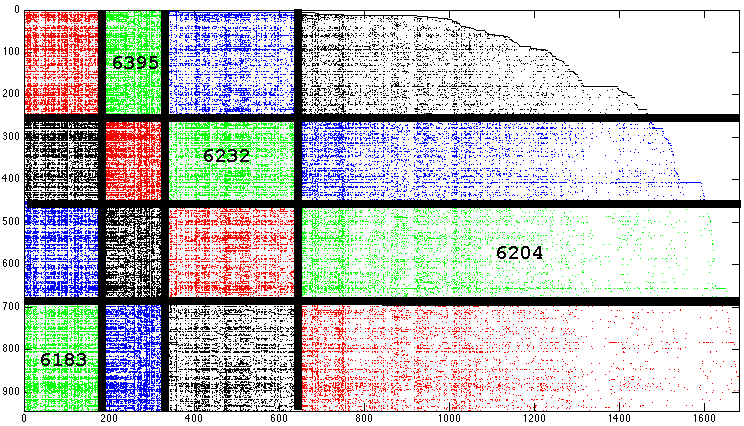} 
      \label{fig:stratb}
    }
  \end{center}
  \caption{Two stratifications for the MovieLens 100K matrix that
    contains 100K ratings from 943 users (rows) on 1682 movies
    (columns). (Left) \STR uses equi-length intervals as suggested in
    the literature. (Right) \BSTR uses a greedy balancing heuristic
    with non-equal intervals. The numbers on the rectangles show the
    number of non-zeros/functions within that block.}
  \label{fig:stratall}
\end{figure}

The third scheme we experimented is {\em stratification} that has
previously been applied to parallelize stochastic gradient descent in
a distributed-memory setting~\citep{gemulla2011}. When the functions
and/or observations are in a structured, multi-dimensional form, such
as a matrix or a tensor, stratification partitions the (parameter)
dimensions into a number of intervals and form a number of function
strata that are amenable to parallelization. Figure~\ref{fig:stratall}
shows two possible stratifications for the MovieLens 100K matrix that
contains 100K ratings from 943 users (rows) on 1682 movies
(columns). The functions in different strata are colored with
different colors: each stratum in the figure corresponds to a subset
$S_k$ for $k \in \{1, 2, 3, 4\}$ and each small, colored rectangle
corresponds to a block; \textit{i.e.}, $B_k = 4$ for $k \in \{1, 2, 3,
4\}$.
Since the block intervals are different within each subset, the blocks
of the same subset $S_k$ can be processed in parallel by $B_k$
threads. In the literature, equi-length intervals~(formed by the long
thick lines in the figure) have been used for stratification and we
call this variant as \STR. The original scheme by~\citet{gemulla2011}
does not consider the balance on the number of functions within the
blocks of the same subset.

As Figure \ref{fig:strat} shows, for \STR, the sparsity and
irregularity of the datasets can yield an unbalanced function-to-block
distribution, which implies a load imbalance for a parallel
execution. For this reason, we propose a balanced stratification
scheme, \BSTR, which aims to generate blocks that contain (almost) the
same number of functions for each subset. For a single dimension, the
intervals are generated as follows: Starting from an empty interval,
the current interval is extended by one as long as it contains less
than $|\cI| / K$ functions. If the number of functions surpasses this
value, the current interval is finalized and the next one is
started. This process is repeated for each dimension to generate the
blocks. This greedy procedure does not consider all the dimensions at
once, and hence, it does not directly balance the number of functions
within the rectangle block. However, according to our experiments, it
performs much better than the original \STR scheme. For instance, for
MovieLens 100K, the second block $S_2$ of a 4-way \STR contains blocks
with 6624, 3492, 2449, and 13585 functions. As Figure \ref{fig:stratb}
shows, these values for \BSTR are 6395, 6232, 6204 and 6183,
respectively.

\begin{algorithm}[htbp]
 \DontPrintSemicolon
 \SetKwInOut{Input}{input}
 \Input{$x$, \texttt{schedule}, $\eta$, $\gamma$ and $\balpha_{k,b}$ for all
   $k = 1, \cdots, K$; $b = 1, \cdots, B_k$.}
 $t = 1$, $S = \mathbf{0}$,  $Y = \mathbf{0}$\;
 ${\CK} = [K, 1, 2, \cdots, K-1, K]$ \;
 \Repeat{convergence or $t$ > max\_epochs}{
    $\beta_t = (\eta t)^\gamma$ \;
    $\CK \leftarrow$ {\sc SetSchedule}($\CK$, \texttt{schedule}) \;
    \For{$\ell = 1, 2, \cdots, K+1$} { \label{ln:lbfgs:for}
    $k = {\CK}[\ell]$\;
    \ForPar{$b=1, 2, \cdots, B_k$}{
       $g_{\balpha_{k,b}} = \nabla f_{k,b}(x_{\balpha_{k,b}})$\; \label{ln:lbfgs:grad2}
   	  \If{$\ell = 1$} {
    		$\bar{g}_{\balpha_{k,b}} = g_{\balpha_{k,b}}$, $\bar{x}_{\balpha_{k,b}} = x_{\balpha_{k,b}}$\; \label{ln:prevs}
   	 }
          $x_{\balpha_{k,b}} = x_{\balpha_{k,b}} - \frac{1}{\beta_t} (\sigma g_{\balpha_{k,b}} + W_{\balpha_{k,b}}NW_{\balpha_{k,b}}\tr g_{\balpha_{k,b}})$\; \label{ln:lbfgs:update}
    } 

    } 
    $s = x - \bar{x}$, 
    $y =g - \bar{g}$, \; \label{ln:lbfgs:hstart}
    \If{$s\tr y > 0$}{
    $\sigma = \frac{s\tr y}{y \tr y}$ \;
    $S = [S(:,2:M),  s]$, \ $Y = [Y(:,2:M), y]$, \ $W = [S, \
    \sigma Y]$\; \label{ln:lbfgs:memory}
    $C = Y\tr Y$, \ $R = \mbox{triu}(S\tr Y)$, \ $D = \mbox{diag}(R) I$\; 
    $N = \left[\begin{array}{cc}
                 R\invt(D + \sigma C) R\inv & -R\invt \\
                 -R\inv & 0
                 \end{array}\right]$\;\label{ln:lbfgs:hend}
    }
    $t \leftarrow t + 1$\;
 }
 \caption{HAMSI with L-BFGS Updates} 
  \label{algo:hlbfgs}
 \end{algorithm}

\section{Example Implementation}
\label{sec:impl}

We give a particular implementation of HAMSI in
Algorithm~\ref{algo:hlbfgs} for illustrative purposes. Here, the
approximate Hessian matrices are obtained using the BFGS quasi-Newton
update formula.  In particular, the compact form of limited memory
BFGS (L-BFGS) is used in inner iterations to form the quadratic models
and obtain their analytical solutions \citep{byrd:1994}. L-BFGS allows
the computation of $(H^{(t)}+\beta_t I)^{-1}v$ for a given vector $v$
without forming any $|\cJ| \times |\cJ|$ matrices, and without any
$\mathcal{O}(|\cJ|^2)$ operations. Moreover, the memory requirement is
only $\mathcal{O}(M|\cJ|)$, where $M$ is the memory size. This step
corresponds to the exact solution of our subproblem in HAMSI (line
\ref{ln:update2} of Algorithm \ref{algo:hamsi}).

Quasi-Newton approximations require the difference of two consecutive
iterates as well as the difference of the gradients evaluated at those
iterates. In line \ref{ln:prevs} of Algorithm~\ref{algo:hlbfgs}, we
store the previous iterate as well as its gradient. Then, in line
\ref{ln:lbfgs:hstart} the desired differences $s$ and $y$ are
evaluated after the new iterate, $x$ and its gradient, $g$ are
evaluated. A limited memory quasi-Newton algorithm uses a collection
of $M$ such differences to update the approximate Hessian; we denote
the corresponding memory matrices of size $|\cJ|\times M$ as $S$ and
$Y$. The vector and matrix algebra operations from line
\ref{ln:lbfgs:hstart} to line \ref{ln:lbfgs:hend} are nothing else but
the direct use of compact form formulas; we refer to \cite{byrd:1994}
for details.

In Algorithm \ref{algo:hamsi}, the {\sc SetSchedule} function rotates over
all the subsets in the same cyclic order. However, in the literature, it is
also quite common to select the subsets randomly without replacement;
\textit{cf.} \citep{gemulla2011}
In the next section, we shall compare our results against the random
order. Therefore, in our example implementation we have also added one
more parameter to the {\sc SetSchedule} function. The parameter
\texttt{schedule} can be either \textit{deterministic} (\texttt{det})
or \textit{stochastic} (\texttt{stoc}). The deterministic order is
the one cyclic left-shift as in Algorithm \ref{algo:hamsi}. The
stochastic order, on the other hand, obtains a temporary array by a
random permutation of the first $K$ elements of $\CK$ and then,
appends this temporary array with the first element in the array. The
resulting array then becomes the new $\CK$ in line \ref{ln:setK}. An
illustrative execution of the new \textsc{SetSchedule} function for
$K=4$ is as follows:
\begin{center}
  \begin{tabular}{ccc}
    Iteration & Deterministic & Stochastic \\
    $t$ & $\CK = [3, 4, 1, 2, 3]$ & $\CK = [3, 1, 4, 2, 3]$\\
    $t+1$ & $\CK = [4, 1, 2, 3, 4]$ & $\CK = [2, 4, 3, 1, 2]$\\
    $t+2$ & $\CK = [1, 2, 3, 4, 1]$ & $\CK = [3, 2, 1, 4, 3]$ \\
    $\vdots$ & $\vdots$ & $\vdots$
  \end{tabular}
\end{center}

All the vector and matrix operations~(including the memory copy
operations) in Algorithm~\ref{algo:hamsi} are parallelized; as stated,
the gradient is computed in parallel by using the techniques in
Section~\ref{sec:par}. For practical purposes, while processing the
blocks of a subset $S_k$, we postpone the partial iterate updates at
line~\ref{ln:lbfgs:update} and do a single lazy update
$x = x - \frac{1}{\beta_t} (\sigma g + WNW\tr g)$ once for each
iteration of the for loop at line~\ref{ln:lbfgs:for}. Furthermore, for
memory and cost efficiency, this update is performed (in parallel)
from right-to-left; \textit{i.e}., following the order in
$W(N(W\tr g))$. In this way, we do not need to store $WNW\tr$ and
perform three matrix-vector multiplications instead of expensive
matrix-matrix multiplications. These operations are parallelized by
block-partitioning of the columns and the rows of $W\tr$ and $W$,
respectively. In addition, the (potentially expensive) shift
operations for the matrices $S$ and $Y$~(at
line~\ref{ln:lbfgs:memory}) are avoided by inserting the new $s$ and
$y$ vectors on top of an existing column in a round-robin fashion and
only updating the corresponding row/column of
$R = \mbox{triu}(S\tr Y)$ at every outer iteration.

\section{Computational Study}
\label{sec:compstud}

All the simulation experiments in this section are performed on a
single machine running on 64 bit CentOS 6.5 equipped with 384GB RAM
and a dual-socket Intel Xeon E7-4870 v2 clocked at 2.30 GHz, where each
socket has 15 cores (30 in total). Each core has a 32kB L1 and a 256kB
L2 cache, and each socket has a 30MB L3 cache. All the codes are
compiled with {\tt gcc 4.9.2} with the {\tt -O3} optimization flag
enabled. For parallelization, we used {\tt OpenMP} and for matrix
operations, we use {\tt GSL v1.13} and compile the codes with options
{\tt -lgsl -lgslcblas}.

\renewcommand{\arraystretch}{1}
\begin{table}[htbp]
\setlength{\tabcolsep}{4pt}
\small
\begin{center}
\scalebox{0.95}{
\begin{tabular}{|c||c|c||ccccc|}
\hline
&  & &\multicolumn{5}{c|}{Average Final RMSE Value}\\
Dataset& Algorithm & \texttt{schedule} &\HW&   \CLR&    \BCLR&       \STR&       \BSTR\\\hline
\multirow{2}{*}{\scriptsize{$\underset{\text{ratings}}{1M}$ -- $\underset{\text{users}}{6040}$- -- $\underset{\text{movies}}{3883}$}}	&	\multirow{2}{*}{mb-GD} &	\texttt{det} & 3.1074	&	3.1061	&	3.0845	&	2.5315	&	2.4588	\\	
	&		& \texttt{stoc} &	3.1433	&	3.1470	&	3.1003	&	2.5325	&	2.4650	\\	\cline{2-8}
\multirow{1}{*}{\scriptsize{(25 seconds)}}	&	\multirow{2}{*}{HAMSI}	& \texttt{det} &	0.6901	&	0.6955	&	0.7102	&	0.6133	&	0.6022	\\
	&		&\texttt{stoc}&	0.6900	&	0.7987	&	0.8017	&	0.6088	&	0.5994	\\	\hline
\multirow{2}{*}{\scriptsize{$\underset{\text{ratings}}{10M}$ -- $\underset{\text{users}}{71567}$ -- $\underset{\text{movies}}{10681}$}}	&	\multirow{2}{*}{mb-GD}	&\texttt{det} &	4.3167	&	4.2676	&	4.2617	&	4.0029	&	3.4088	\\	
	&		&\texttt{stoc}&	4.3009	&	4.2863	&	4.2801	&	4.0035	&	3.4094	\\	\cline{2-8}
\multirow{1}{*}{\scriptsize{(250 seconds)}}	&	\multirow{2}{*}{HAMSI}	& \texttt{det}  &	0.9279	&	1.0181	&	0.8941	&	0.8923	&	0.8643	\\
	&		&\texttt{stoc}&	0.9207	&	1.1357	&	1.1229	&	0.8988	&	0.8652	\\	\hline
\multirow{2}{*}{\scriptsize{$\underset{\text{ratings}}{20M}$ -- $\underset{\text{users}}{138493}$ -- $\underset{\text{movies}}{26744}$}}	&	\multirow{2}{*}{mb-GD}	& \texttt{det}  &	4.8655	&	4.8051	&	4.8000	&	4.8093	&	3.8890	\\	
	&		&\texttt{stoc}&	4.8641	&	4.8279	&	4.8142	&	4.8091	&	3.8975	\\	\cline{2-8}
\multirow{1}{*}{\scriptsize{(500 seconds)}}	&	\multirow{2}{*}{HAMSI}	& \texttt{det} &	1.0170	&	1.1117	&	0.9521	&	1.0113	&	0.9042	\\
	&		&\texttt{stoc}&	1.0112	&	1.2944	&	1.2220	&	1.0231	&	0.9035	\\	\hline
\end{tabular}
}
\end{center}
\caption{Root-mean square errors of all parallel execution strategies
  for mini-batch GD and HAMSI for three datasets. For each
  configuration, both algorithms are given the same initial solution,
  run for 25, 250, and 500 seconds on 1M, 10M and 20M rating datasets,
  respectively, with 16 threads. This experiment is performed three
  times with three different initial solutions and the average is
  reported.}
\label{tab:main}
\end{table}

\subsection{Setup}
For the experiments, we use the MovieLens datasets with 1M, 10M, and
20M ratings \citep{Harper:2015}. The number of entities (users and
movies) for each dataset are given in Table~\ref{tab:main}. For each
dataset, we form the corresponding User $\times$ Movie matrix and
compute a factorization of rank $50$; \textit{i.e}., the latent
(inner) dimension of the factors is set to $50$. To compare the
performance of HAMSI, we implemented the mini-batch gradient
descent~(mb-GD) as a baseline. In particular, both HAMSI and mb-GD use
the same code base with minor differences: mb-GD uses a different
schedule order $\CK$ of size $K$ (instead of $K+1$), which is simply a
permutation of $\{1,2, \cdots, K\}$ and the stochastic gradient update
equation $x \leftarrow x - \frac{1}{\beta_t}g$ instead of the one
altered with the approximate Hessian (line~\ref{ln:lbfgs:update}). We
note that in the first few iterations, until full $S$ and $Y$ matrices
are obtained, HAMSI also employs the gradient descent update. Clearly,
mb-GD implementation does not perform any of the matrix operations at
lines~\ref{ln:lbfgs:hstart}--\ref{ln:lbfgs:hend}. The constant
algorithm parameters are set as follows: for mb-GD,
$(\eta=0.001, \gamma=0.51)$; for HAMSI, $(\eta=0.06, \gamma=0.51)$.

We carried out two sets of experiments: (I) fixed wallclock time and
(II) fixed number of iterations.  In the first set (I), we allocate a
fixed optimization time for each algorithm and let the code run for
$25$, $250$ and $500$ seconds for the datasets 1M, 10M and 20M,
respectively. We report the training error, where we measure the
quality of the final solution by the root-mean-square error~(RMSE).
We employed all 5 parallelization schemes, \HW, \CLR, \BCLR, \STR, and
\BSTR with $1, 2, 4, 8$ and $16$ threads, and repeated each experiment
three times with different random seeds used to generate the initial
solutions. Overall, 3~(datasets) $\times$ 2~(algorithms) $\times$
5~(parallelization scheme) $\times$ 5~($\#$threads) $\times$ 3~(random
seeds) $\times$ 2~(scheduling) $= 900$ experiments are performed. For
this set of experiments, the schemes \HW, \CLR and \BCLR uses $K = 20$
where \STR and \BSTR automatically sets $K$ to number of threads. As
explained in Section~\ref{sec:par}, for \CLR and \BCLR, we packed the
color sets into $K = 20$ large batches, since there were around $3500$,
$35000$, and $65000$ color sets for the datasets 1M, 10M, and 20M
respectively.

Table~\ref{tab:main} shows the RMSE averages for parallel executions
with 16 threads for the first experiment set (I). Since the algorithms
are executed for a fixed amount of time but not for a fixed number of
iterations, the parallelization scheme has a significant impact on the
overall optimization performance. For both of the algorithms, \BSTR
with a deterministic schedule is the best candidate in almost all the
experiments. Furthermore, for HAMSI, the parallelization scheme \BCLR
is clearly better than \CLR. In general, HAMSI outperforms mini-batch
GD in terms of the final RMSE value. Since both HAMSI and mb-GD use
the same codebase, and mb-GD is allowed to perform more iterations in
a fixed amount of time, we conclude that it pays well off doing the
extra computations for the Hessian approximations overhead in HAMSI.

\begin{figure}[t]
\centering
\includegraphics[width=1.00\columnwidth]{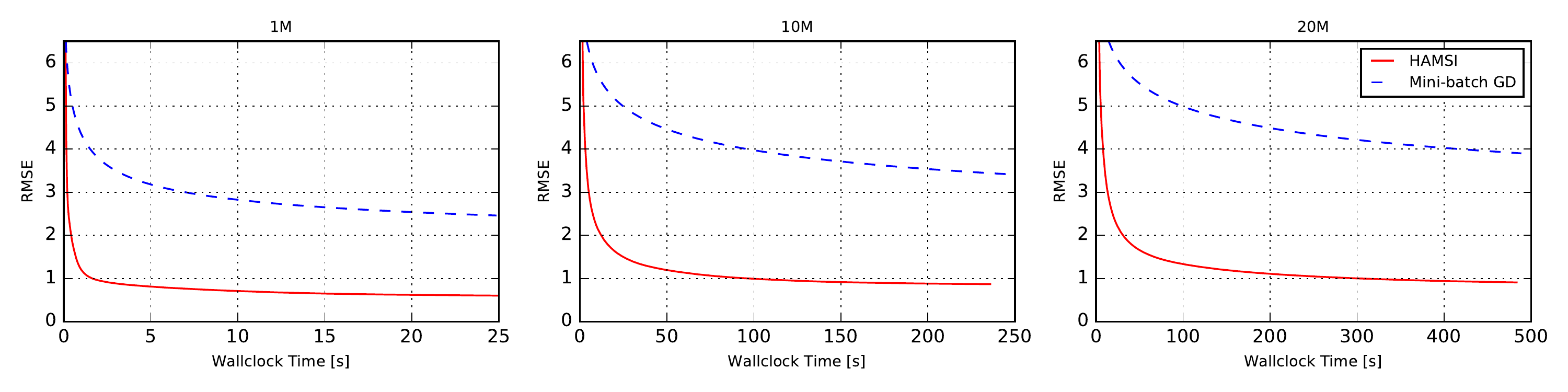}
\vspace*{-4ex}
\caption{Convergence of mb-GD and HAMSI in terms RMSE values with 16
  threads.}
\label{fig:cmp_best_rmse}
\end{figure}

To compare the algorithms further, we select the best performing
parameter settings from Table~\ref{tab:main}; that is, \BSTR scheme
with deterministic scheduling. Figure~\ref{fig:cmp_best_rmse} shows
the progress of these mb-GD and HAMSI variants in terms of RMSE
values. Clearly, HAMSI drastically improves the RMSE value for the
corresponding problem much earlier than mb-GD. Furthermore, the
differences in the attained RMSE values seem to increase with the size
of the problem. This result also illustrates the performance gain
obtained by using the second order information.

\begin{figure}[t]
\centering
\includegraphics[width=1.00\columnwidth]{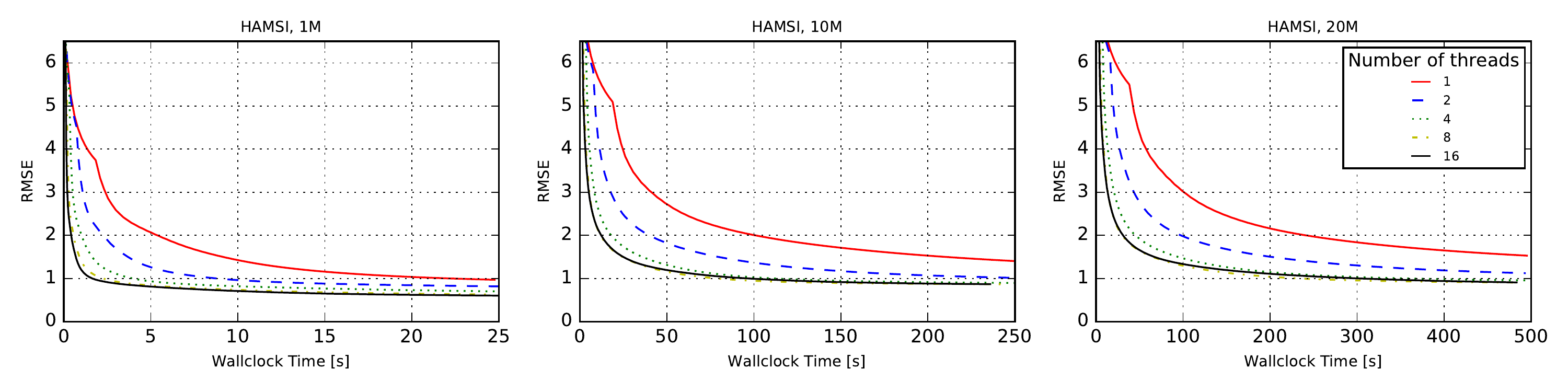}
  \vspace*{-4ex}
  \caption{Convergence behaviors of HAMSI when the number of threads
    is increased.}
  \label{fig:cmp_best_threads_hamsi}
\end{figure}

Our next discussion is about the changes in the convergence behavior
of HAMSI when the number of threads is altered. \BSTR is used with a
$K$ that is equal to the number of threads. Thus, the batch size
becomes smaller when the number of threads increases. Consequently,
the estimate of the original gradient deteriorates during the
incremental steps leading to a somewhat high variance. However, Figure
\ref{fig:cmp_best_threads_hamsi} shows that the performance of HAMSI
is not affected significantly by the high variance gradient
estimator. On the contrary, HAMSI seems to converge quickly for all
problems when the number of threads is increased. This point is
complementary to our observations in the next subsection, where we
have demonstrated that HAMSI scales nicely with more computational
resources.

\subsection{Efficiency of Parallelization}
\label{subsec:efficiency}

To further analyze the efficiency of the parallelization schemes and
understand the differences in the final RMSE values, we partition the
cost of an outer iteration into three phases: (i) data pass and
gradient computation~(line~\ref{ln:lbfgs:grad2} of
Algorithm~\ref{algo:hlbfgs}); (ii) performing the
updates~(line~\ref{ln:lbfgs:update} of Algorithm~\ref{algo:hlbfgs});
(iii) maintaining the approximate
Hessian~(lines~\ref{ln:lbfgs:hstart}--\ref{ln:lbfgs:hend} of
Algorithm~\ref{algo:hlbfgs}).

\begin{figure}[t]
  \begin{center}
  \subfigure[1M]{\includegraphics[width=0.72\columnwidth]{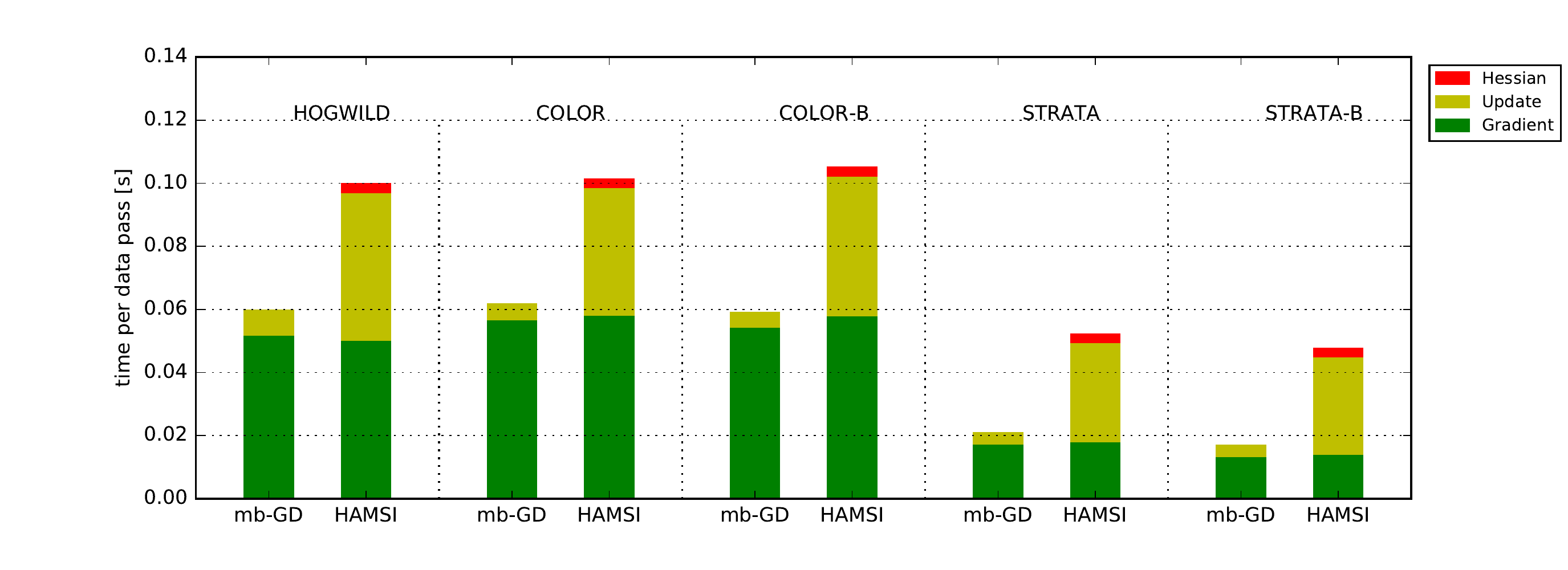} \label{fig:spd:1M}}
   \subfigure[10M]{\includegraphics[width=0.72\columnwidth]{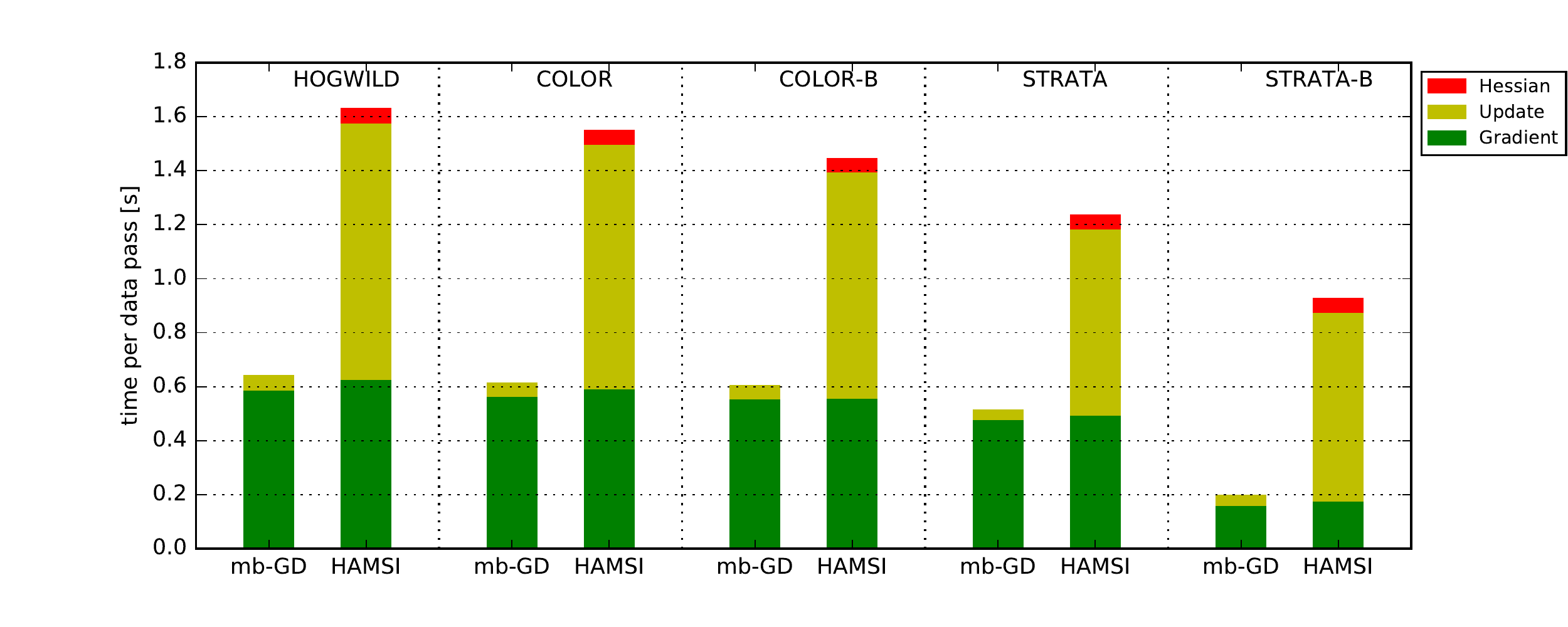} \label{fig:spd:10M}}
    \subfigure[20M]{\includegraphics[width=0.72\columnwidth]{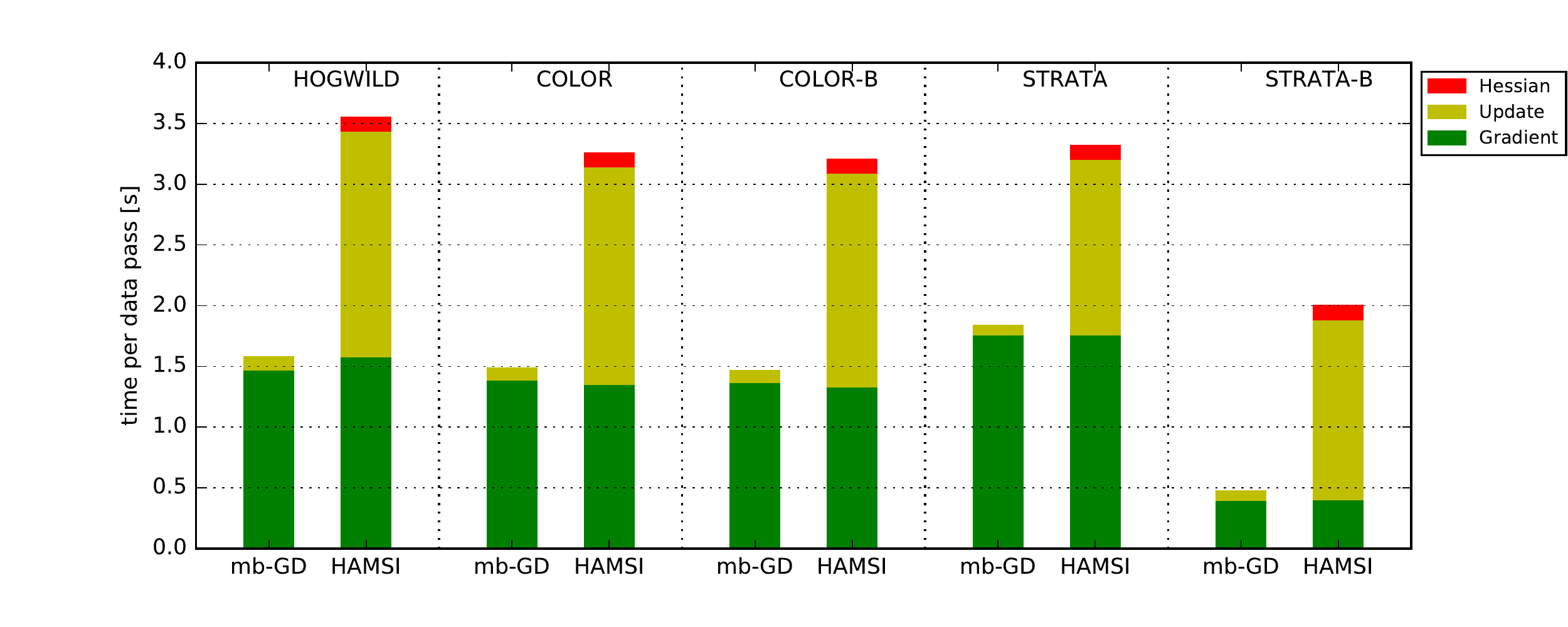} \label{fig:spd:20M}}
  \end{center}
  \caption{Hessian computation, update, and gradient computation time
    for an outer iteration of mb-GD and HAMSI with 16 threads.}
  \label{fig:threephase}
\end{figure} 

Figure~\ref{fig:threephase} shows the times spent for these three
phases with 16 threads. With HAMSI, the Hessian approximation is much
cheaper than the other two phases. With mb-GD, computing the gradient,
not surprisingly, is the main bottleneck. Hence, the parallelization
schemes proposed in this work can be used to make a
gradient-descent-based algorithm faster, and hence, more effective in
practice. As mentioned above, although mb-GD performs $2$-$4\times$
more iterations per experiment, as Table~\ref{tab:main} and
Figure~\ref{fig:cmp_best_rmse} show, HAMSI's final RMSE values are
much better than that of mb-GD's.

Among all the parallelization schemes, \BSTR is the best one in terms
of computational efficiency. Indeed, \BSTR is expected to be better
than \STR, since the load is distributed to the threads more
evenly. In addition, with \BSTR, each thread accesses a
specific/narrowed part of $x$ and $g$. However, with \HW and coloring
based schemes, the $x$ and $g$ locations accessed by a single thread,
are distributed among all $x$ and $g$. Hence, stratification is
expected to be a more cache-friendly parallelization scheme. To verify
this, we performed a second set of experiments with 16 threads for
which, each HAMSI variant, which employs a different parallelization
scheme, is run for $100$ (outer) iterations and the cache miss ratios
are observed with the {\tt perf} command. To perform the same number
of gradient computations, we keep $K = 16$ for all the parallelization
schemes in this experiment. Furthermore, by taking the
function-to-thread assignment into account for each scheme and
assuming that each function takes a unit time to process, we computed
the total parallel gradient computation work for an outer iteration
which is equal to
\[
parWork = \sum_{k = 1}^K{\lceil |S_k| / 16 \rceil} \mbox{\ \ \ \ and\
  \ \ \  } parWork = \sum_{k = 1}^K{\max_{b \in \{1, \cdots,
    B_k\}}\left( |S_{k,b}|\right)}.
\]
The former one is for \HW, \CLR and \BCLR, and the latter one is for
\STR and \BSTR. The experiments are performed three times with three
different initial solutions. Table~\ref{tab:cache} shows the averages
of these experiments.

\begin{figure}[t]
  \begin{center}
  \subfigure[1M]{\includegraphics[width=0.52\columnwidth]{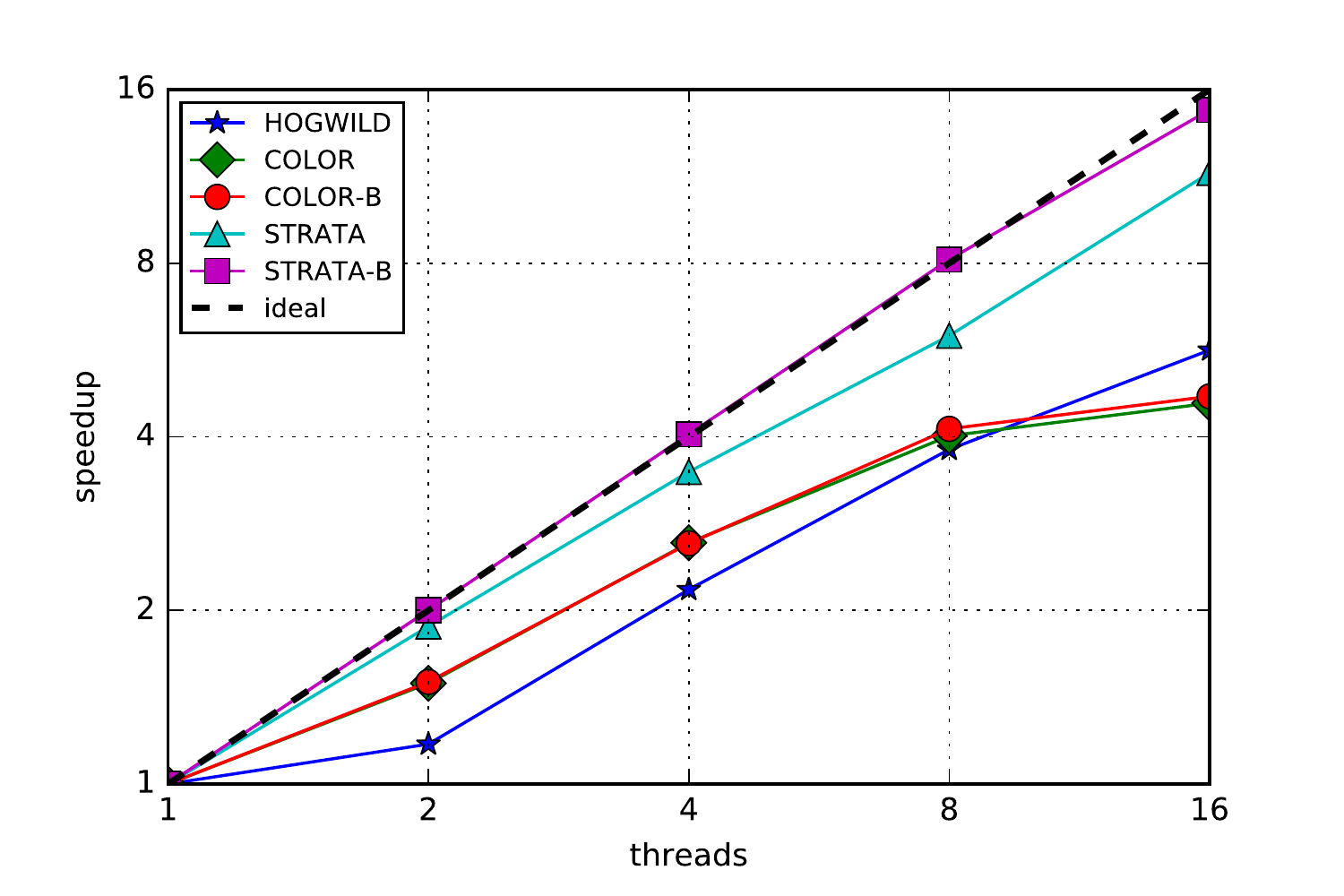} \label{fig:spd2:1M}}
   \subfigure[10M]{\includegraphics[width=0.52\columnwidth]{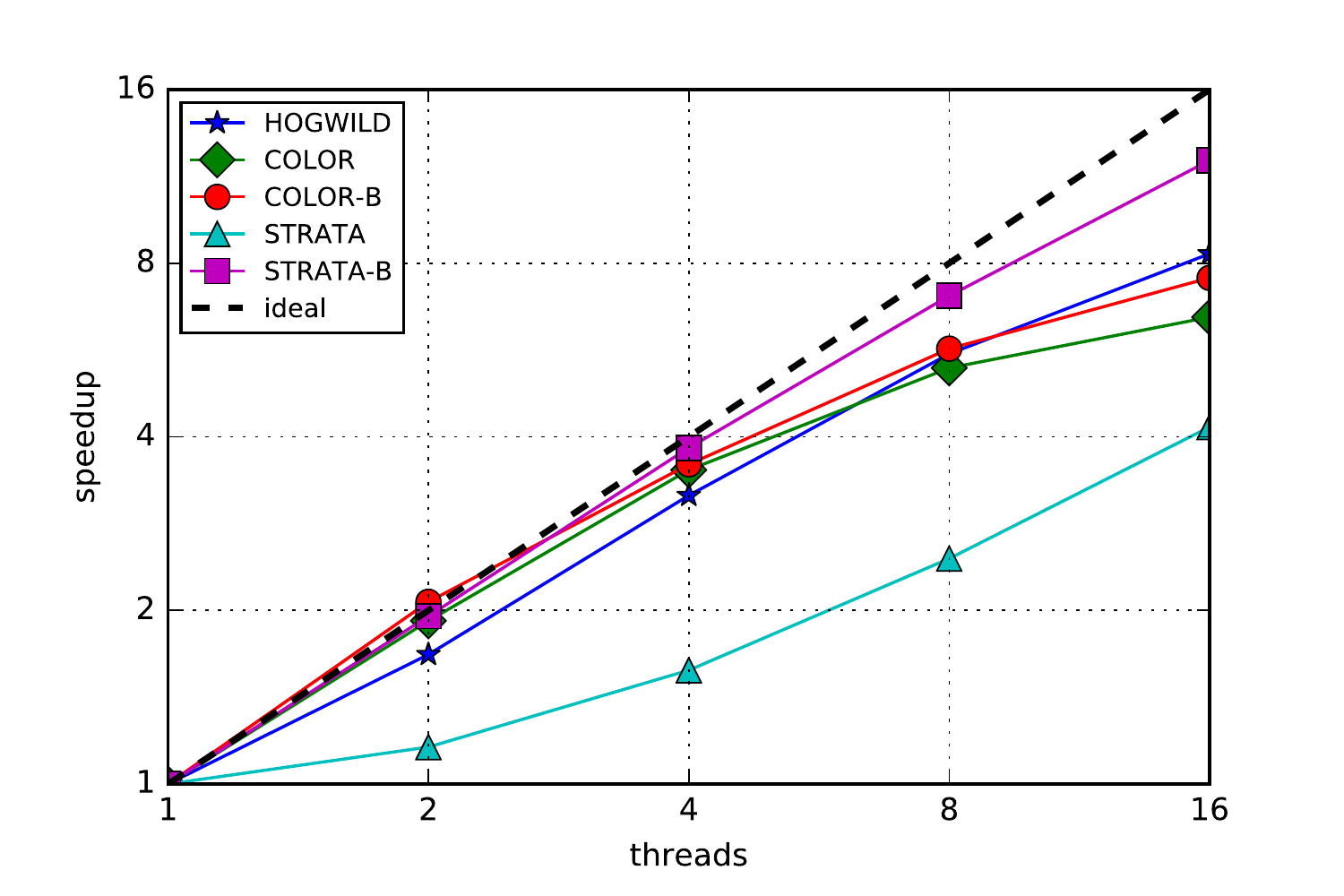} \label{fig:spd2:10M}}
    \subfigure[20M]{\includegraphics[width=0.52\columnwidth]{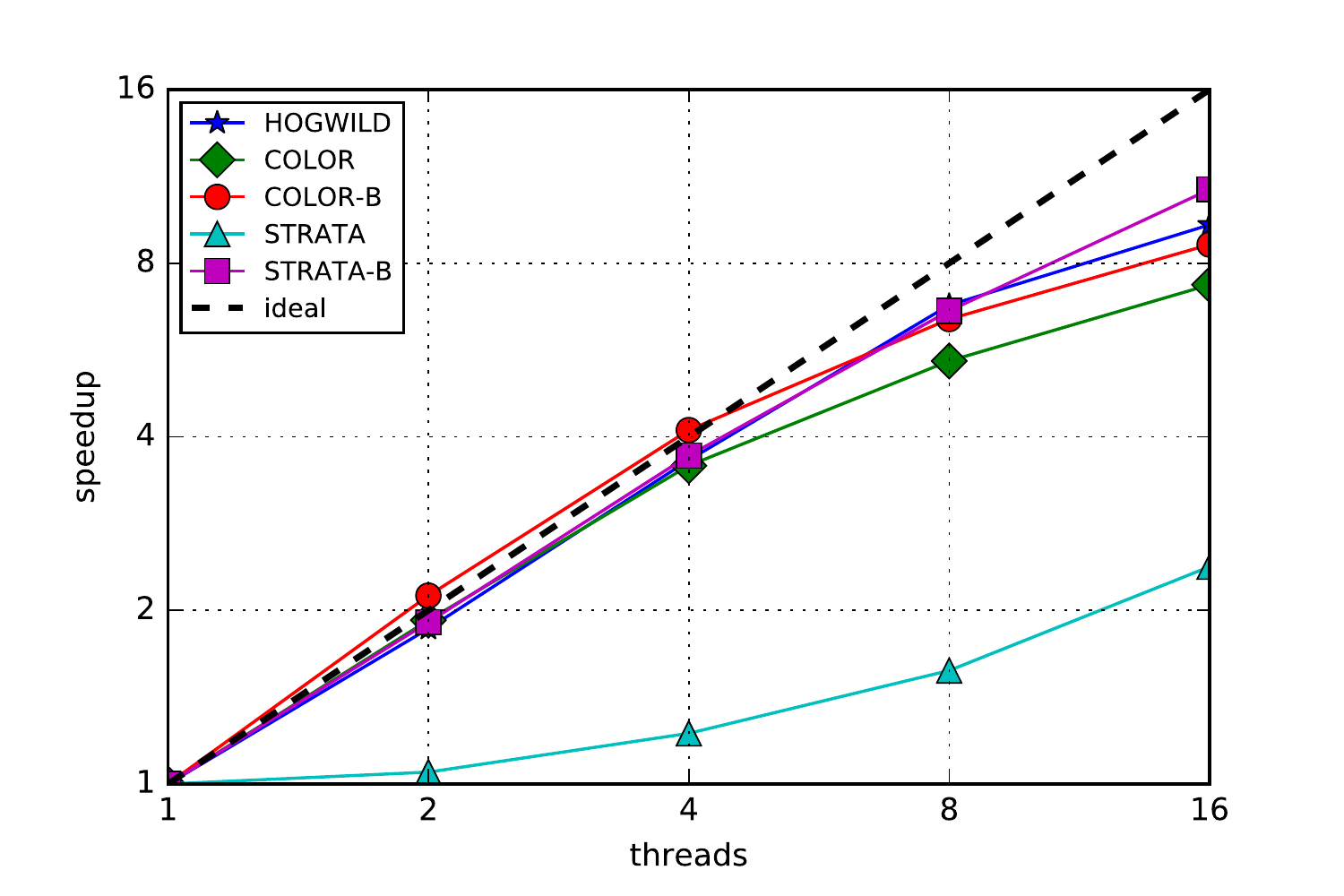} \label{fig:spd2:20M}}
  \end{center}
  \caption{Speedups obtained on the gradient computation phase for
    each parallelization scheme with $1$, $2$, $4$, $8$ and $16$
    threads for 1M, 10M, and 20M rating datasets.}
  \label{fig:speedup}
\end{figure}

As expected, although \HW perfectly balances the load, it has the
worst cache-miss ratio average. Furthermore, the stratification-based
schemes are more cache friendly compared to others. Among them, \STR
suffers due to a high $parWork$ value especially for $10M$ and $20M$
rating datasets and it is one of the slowest. On the other hand, \BSTR
is the fastest, since it also solves the load balancing problem thanks
to the proposed greedy approach to create balanced blocks. The
performance difference between \BSTR and \HW~(as well as \CLR and
\BCLR) is significant since the cost of a cache access increase $3$ to
$4$ times for each cache level\footnote{\textit{cf.}
  http://www.7-cpu.com/}. Moreover, the impact of being cache friendly
will be more important with increasing number of threads and larger
datasets. Figure~\ref{fig:speedup} shows the speedups for the gradient
computation phase of HAMSI for all the datasets: \BSTR, as expected,
is more scalable than the other schemes investigated in this work.

There exist other studies in the literature that aims providing load imbalance and
good cache utilization at the same time. For 
instance, recently \citet{NIPS2016_6604} proposed {\sc Cyclades} that samples and uses small, connected
subgraphs of the function graph. They compared the performance of
the proposed method with that of \HW on SGD with the 10M rating dataset. 
Connectivity implies cache locality up to some extent; indeed, a set of connected functions
touch common factors. However, the corresponding subgraph of the factor graph
may not be well-connected, where the connectedness, which is correlated with data reuse, can be quantified with clustering coefficients, edge-to-vertex ratio, etc. 
Furthermore, without arranging the locations of the factors on memory, e.g., via relabeling and modifying the input in a preprocessing phase, 
it is not possible to exploit the full potential of the connectivity. For instance, Pan~et~al. reported that with 16 threads, their partitioining phase takes as much time as 5.5 full data passes, i.e., outer iterations, of
\HW where a single \HW data pass takes 0.48 seconds; with {\sc Cyclades} the runtime reduces to 0.42 seconds. 
Although HAMSI is a second-order incremental algorithm and hence not exactly comparable with {\sc Cyclades}, looking to mb-GD~(which is also not directly comparable but the closest one we have), a full data pass takes 0.63 seconds for \HW on 10M rating dataset. 
Note that {\sc Cyclades} experiments are performed on an 18-core Intel Xeon E7-8870 v3 with 576kB L1, 4.5MB L2, and 45MB L3 caches~(32kB, 256kB, and 30MB, respectively, for our processor). Hence, 
due to possible implementation differences and a significantly better cache hierarchy, the runtime for \HW is better. However, the \BSTR parallelization of mb-GD reports 0.2 seconds for a full data pass 
on the same dataset which is 3$\times$ better than \HW where the improvement is around $15\%$ for {\sc Cyclades}. This is expected since \BSTR respects the initial data labels, and hence, how the factor data is organized in the memory.
Furthermore, since it only employs a cheap load-balancing heuristic, \BSTR's preprocessing phase, performed only once, is 50$\times$ faster than a single data pass of \HW. 

\begin{table}
\setlength{\tabcolsep}{4pt}
\begin{center}
\scalebox{0.8}{
\begin{tabular}{|l|rrrr|r|r|r|}
\hline\hline
\multicolumn{8}{|c|}{1M}\\\hline\hline
Scheme	&	L1-dL ($\times 10^9$) 	&	L1-m ($\%$)	&	L2-m  ($\%$)		&	L3-m  ($\%$)		&	$parWork$	&	RMSE	&	Time (sec)	\\ \hline 
\HW	&	75.0	&	13.1	&	8.7	&	1.7	&	62513	&	0.82	&	8.7	\\
\CLR	&	74.8	&	11.8	&	5.5	&	1.1	&	64354	&	0.82	&	8.4	\\
\BCLR	&	75.0	&	11.9	&	5.6	&	1.2	&	64193	&	0.82	&	8.6	\\
\STR	&	74.5	&	8.8	&	2.3	&	0.4	&	98087	&	0.83	&	4.8	\\
\BSTR	&	74.3	&	8.8	&	2.3	&	0.4	&	69145	&	0.82	&	4.3	\\ \hline \hline
\multicolumn{8}{|c|}{10M}\\\hline\hline												
Scheme	&	L1-dL ($\times 10^9$) 	&	L1-m ($\%$)	&	L2-m  ($\%$)		&	L3-m  ($\%$)		&	$parWork$	&	RMSE	&	Time (sec)	\\ \hline 
\HW	&	666.7	&	13.2	&	10.2	&	3.0	&	625003	&	1.03	&	136.5	\\
\CLR	&	673.6	&	11.8	&	7.6	&	1.6	&	642486	&	1.20	&	124.9	\\
\BCLR	&	672.6	&	11.9	&	7.7	&	1.7	&	641484	&	0.97	&	122.1	\\
\STR	&	673.0	&	8.8	&	3.9	&	0.9	&	2535881	&	1.03	&	128.9	\\
\BSTR	&	672.0	&	8.6	&	3.7	&	0.9	&	659763	&	1.01	&	95.1	\\ \hline \hline
\multicolumn{8}{|c|}{20M}\\\hline\hline																									
Scheme	&	L1-dL ($\times 10^9$) 	&	L1-m ($\%$)	&	L2-m  ($\%$)		&	L3-m  ($\%$)		&	$parWork$	&	RMSE	&	Time (sec)	\\ \hline 
\HW	&	1328.5	&	13.3	&	10.5	&	3.9	&	1250016	&	1.12	&	304.6	\\
\CLR	&	1354.9	&	11.7	&	7.9	&	2.1	&	1284808	&	1.28	&	269.8	\\
\BCLR	&	1381.2	&	11.7	&	8.0	&	2.2	&	1281179	&	1.05	&	270.3	\\
\STR	&	1331.8	&	9.0	&	5.0	&	1.1	&	8726743	&	1.13	&	326.8	\\
\BSTR	&	1331.7	&	8.8	&	4.2	&	1.1	&	1289630	&	1.12	&	189.1	\\ \hline \hline
\end{tabular}
}
\end{center}
\caption{Cache miss ratios, $parWork$ and RMSE values, and the
  execution times of HAMSI variants when different parallelization
  schemes are employed with $K = 16$ and 16 threads for the datasets
  with 1M, 10M and 20M ratings. The first column is the number of
  L1-cache data load requests performed by the execution, and the next
  three columns are percentages with respect to the first column.} 
\label{tab:cache}
\end{table}

\section{Conclusion and Future Work}

In this work, we propose HAMSI, Hessian Approximated Multiple Subsets
Iteration, a provably convergent incremental algorithm for solving
large-scale partially separable optimization problems that frequently
emerge in machine learning and demonstrate its use in a matrix
factorization problem with missing values.  HAMSI may be considered as
a viable alternative in many scenarios in practice as we empirically
show in our matrix-factorization experiments. In the near future, we
are planning to adopt the proposed optimization techniques as well as
the parallelization schemes to other problems such as logistic
regression, neural network training and tensor factorization.

HAMSI currently solves unconstrained optimization problems. However,
in several applications, like nonnegative matrix factorization, there
could be simple constraints that need to be taken into
account. Although one can devise a straightforward projection step to
maintain feasibility, our current analysis does not guarantee
convergence. Incorporating simple constraints is in our future
research agenda.

An interesting observation from Table~\ref{tab:cache} is that with
fixed number of iterations, \CLR is the worst scheme in terms of RMSE
values while \BCLR is the best one. What differs these schemes from
the others is an implicit high coverage of the gradient entries for
each subset. Indeed, a valid coloring puts independent functions to
the same set. Hence these functions will cover different gradient
locations. However, the traditional first-fit policy employed by \CLR
generates a few massive color sets with a (probably) full coverage and
many small color sets covering only a few positions. Since the
gradient descent variants perform the updates by following only these
{\em touched} positions, the optimization process may be deteriorated
with the restricted moves implied by \CLR. However, with \BCLR, the
problem will mostly be solved. We claim that although \BCLR does not
improve the computational efficiency due to a limited number of cores
on a shared-memory machine, with a fixed number of iterations, its
RMSE value is lower compared to other schemes since its gradient
coverage is much better. Although we believe that this is an
interesting claim needs to be proven, the investigation of such a
claim is beyond the motivation of this study. In the future, we will
analyze the impact of coverage on the optimization efficiency and
investigate various parallelism schemes featuring a better gradient
coverage, load balance, and cache locality at the same time.

\begin{acks}
This work is supported by the Scientific and Technological Research Council of Turkey (TUBITAK) Grant no. 113M492.
\end{acks}


\appendix

\section{Omitted Proofs}
\label{sec:appendix}

\subsection*{Proof of Lemma \ref{lem1}}
By using Assumption A.2, we have
\[
\begin{array}{rl}
\|\nfsx{[l]}{t,l-1} - \nfsx{[l]}{t} \|  & \leq L\|\xit{t,l-1} - \xit{t} \| \\[2mm]
& = L\|\xit{t,l-1} - \xit{t, l-2} + \xit{t, l-2} - \xit{t, l-3} + \cdots + \xit{t,1} - \xit{t}\| \\[2mm]
& \leq L \sum_{j=1}^{\ell-1}\|\xit{t,j} - \xit{t, j-1}\|.
\end{array}
\]
Note for $j=1, \cdots, \ell-1$ that
\[
\|\xit{t,j} - \xit{t, j-1}\| = \|\xit{t,j-1} - (\Hit{t} +
                               \beta_tI)\inv\nfsx{[j]}{t,j-1} -
                               \xit{t, j-1}\| \leq M_t C,
                               \]
where the last inequality holds by Assumption A.4. Therefore, we have
\[
\|\nfsx{[l]}{t,l-1} - \nfsx{[l]}{t} \| \leq LM_tC(\ell - 1).
\]
\subsection*{Proof of Lemma \ref{lem2}}

At iteration $t + 1$, we have
\[
\begin{array}{rl}
\xit{t+1} & = \xit{t} - \sum_{\ell=1}^{K+1} (\Hit{t} +
            \beta_tI)^{-1}\nfsx{[\ell]}{t, \ell-1} \\ [2mm]
          & = \xit{t} - (\Hit{t} + \beta_tI)^{-1}\nabla f(\xit{t}) \\ [2mm]
          & \hspace{5mm} + (\Hit{t} + \beta_tI)^{-1}\sum_{\ell=1}^{K+1} (\nfsx{[\ell]}{t} -
            \nfsx{[\ell]}{t,\ell-1}).
\end{array}
\]
This shows that 
\begin{equation}
    \xit{t+1} - \xit{t} = \Delta_t - (\Hit{t} + \beta_tI)^{-1}\nabla
    f(\xit{t}),
\label{eq:lem2:1}
\end{equation}
where   
\[
\Delta_t\equiv (\Hit{t} + \beta_tI)^{-1}\sum_{\ell=1}^{K+1}
(\nfsx{[\ell]}{t} - \nfsx{[\ell]}{t,\ell-1}).
\]
Using now \eqref{eq:lem1} implies
\begin{equation}
\label{eq:lem2:2}
\begin{array}{rl}
\|\Delta_t\| & \leq M_t\sum_{\ell=1}^{K+1} \|\nfsx{[\ell]}{t} - \nfsx{[\ell]}{t,\ell-1}\| \\ [2mm]
            & \leq M_t\sum_{\ell=1}^{K+1}LM_tC(\ell-1) =
              \frac{1}{2}LM^2_tC K(K+1) = BM_t^2. \\ [2mm]
\end{array}
\end{equation}
Then, we obtain
\[
\begin{array}{rl}
\| \xit{t+1} - \xit{t}\| & = \|\Delta_t - (\Hit{t} + \beta_tI)^{-1}\nabla f(\xit{t})\| \\[2mm]
& \leq \|\Delta_t\| + \|(\Hit{t} + \beta_tI)^{-1}\|\nabla f(\xit{t})\|\\[2mm]
& \leq BM_t^2 + CM_t \leq \frac{B + C(M+1)}{M+1}M_t.
\end{array}
\]

\subsection*{Proof of Theorem \ref{thm:conv1}}

As $f$ is a twice differentiable function, we have
\[
\fx{t+1} - \fx{t} \leq {\nfx{t}}^{T} (\xit{t+1} - \xit{t}) +
\frac{LK}{2} \| \xit{t+1} - \xit{t}\|^2.
\]
Using now Lemma \ref{lem2} along with \eqref{eq:lem2:1} and
\eqref{eq:lem2:2}, we obtain
\begin{equation}
\label{eq:thm:1}
\begin{array}{rl}
  \fx{t+1} - \fx{t} & \leq \nfx{t}^T \Delta_t - \nfx{t}^T (\Hit{t} +
                       \beta_tI)^{-1}\nfx{t} + \frac{LK}{2}\|\xit{t+1}
                      - \xit{t}\|^2 \\ [2mm]
& \leq \|\nfx{t}\| \||\Delta_t\| - U_t\| \nfx{t} \|^2 +
  \frac{LK}{2}\left(\frac{B + C(M+1)}{M+1}\right)^2M^2_t \\ [2mm]
& \leq - U_t\| \nfx{t} \|^2 + \bar{B}M^2_t,
\end{array}
\end{equation}
where
$\bar{B} \equiv CB + \frac{LK}{2}\left(\frac{B +
      C(M+1)}{M+1}\right)^2$.
Due to Assumption A.1, we can write
$\inf_{x \in \RR^n} f(x) = f^* > -\infty$. Thus, we obtain
\[
0 \leq \fx{t+1} - f^* \leq \fx{t} - f^* + \bar{B}M^2_t.
\]
Relation \eqref{cond_beta} and Lemma 2.2 in \citep{mangasarian:1994}
together show that the sequence $\{\fx{t}\}$ converges. By using
\eqref{eq:thm:1}, we further have
\[
\begin{array}{rl}
  \fx{1} - f^* & \geq \fx{1} - \fx{t} = \sum_{j=1}^{t-1}(\fx{t} - \fx{t+1}) \\ [2mm]
               & \geq \sum_{j=1}^{t-1}U_j \| \nfx{j} \|^2 -  \bar{B}\sum_{j=1}^{t-1}M^2_j \\ [2mm]
               & \geq \underset{1 \leq j \leq t-1}{\inf} \| \nfx{j} \|^2 \sum_{j=1}^{t-1}U_j  -  \bar{B}\sum_{j=1}^{t-1}M^2_j.
\end{array}
\]
Let now $t \rightarrow \infty$, then 
\begin{equation}
\label{eq:thm:2}
\fx{1} - f^* \geq \inf_{j \geq 1}\| \nfx{j} \|^2 \sum_{j=1}^{\infty}U_j - \bar{B}\sum_{j=1}^{\infty}M^2_j.
\end{equation}
Using again Assumption A.1 and conditon \eqref{cond_beta}, we obtain
\begin{equation}
\label{eq:thm:3}
\inf_{t \geq 1}\| \nfx{t} \| = 0.
\end{equation}

Now, suppose for contradiction that the
sequence $\{\nfx{t}\}$ does not converge to zero. Then, there exists
an increasing sequence of integers such that for some $\veps > 0$, we
have $\|\nfx{t_\tau}\| \geq \veps$ for all $\tau$.  On the other hand, 
the relation \eqref{eq:thm:3} implies that there exist some
$j > t_\tau$ such that $\|\nfx{j}\| \leq \frac{\veps}{2}$. Let
$j_\tau$ be the least integer for each $\tau$ satisfying these
inequalities. Then, we have
\[
\begin{array}{rl}
  \frac{\veps}{2} & \leq \|\nfx{t_\tau}\| - \|\nfx{j_\tau}\| \\[2mm]
                  & \leq \|\nfx{t_\tau} - \nfx{j_\tau}\| \\[2mm]
                  & \leq LK\|\xit{t_\tau} - \xit{j_\tau}\| \leq LK\frac{B+C(M+1)}{M+1}\sum_{k=t_\tau}^{j_\tau-1} M_k,
\end{array}
\]
where the last inequality follows from Lemma \ref{lem2}.  Since
$0 < M \leq U$, there exists $\zeta \leq \frac{M}{U} \leq 1$ such that
\[
M + \beta_k \geq \zeta U + \beta_k \geq  \zeta (U + \beta_k) \implies
M_k \leq \frac{1}{\zeta} U_k.
\]
Thus, we obtain 
\[
0 < \hat{B} \equiv \frac{\veps(M+1)\zeta}{2LK(B+C(M+1))} \leq \sum_{k=t_\tau}^{j_\tau - 1} U_k.
\]
Then, using together with inequality \eqref{eq:thm:1}, we obtain
\[
\begin{array}{rl}
\fx{t_\tau} - \fx{j_\tau} & \geq \sum_{k=t_\tau}^{j_\tau-1} U_k\|\nfx{k}\|^2 - \bar{B}\sum_{k=t_\tau}^{j_\tau-1}M_k^2 \\[2mm]
                         & \geq \hat{B} \underset{t_\tau \leq k \leq j_\tau-1}{\inf} \|\nfx{k}\|^2 - \bar{B}\sum_{k=t_\tau}^{\infty}M_k^2.
\end{array}
\]
Since the left-hand-side of the inequality converges and the condition
\eqref{cond_beta} holds, we have
\begin{equation}
  \label{eq:thm:4}
  \lim_{\tau \uparrow \infty}~\underset{t_\tau \leq k \leq j_\tau-1}{\inf} \|\nfx{k}\|^2 = 0.
\end{equation}
But our choice of $t_\tau$ and $j_\tau$ guarantees
$\|\nfx{k}\| > \frac{\veps}{2}$ for all $t_\tau \leq k \leq j_\tau$,
and hence, we arrive at a contradiction with
\eqref{eq:thm:4}. Therefore, ${\nfx{t}}$ converges, and with the
continuity of the gradient, we conclude for each accumulation point
$x^*$ of the sequence $\{\xit{t}\}$ that $\nabla f(x^*) = 0$ holds.

\clearpage
\vskip 0.2in

\bibliographystyle{ACM-Reference-Format-Journals}
\bibliography{partensor}

\end{document}